\documentclass{article}

\usepackage{graphicx,amsmath,amssymb}
\usepackage{amssymb, amsmath, amsopn}
\usepackage{times,mathptmx}
\usepackage{enumerate}
 \usepackage{mathptmx}
 \usepackage{helvet}
 \usepackage{courier}
 \usepackage{makeidx}
 \usepackage{multicol}
 \usepackage{caption}
 \usepackage{graphicx}
 \usepackage{caption}
 \usepackage{subfig}
 \usepackage{color}
 \usepackage{array}
 \usepackage{verbatim}
\usepackage[algo2e,linesnumbered,vlined,ruled]{algorithm2e}
\usepackage{balance}
\usepackage{amsthm}
\usepackage{dsfont}
\usepackage{algorithmic}
\usepackage[english]{babel}
\usepackage{booktabs}
\usepackage{multirow}

\begin{document}

\newtheorem{observation}{Observation}
\newtheorem{theorem}{Theorem}[section]
\newtheorem{lemma}[theorem]{Lemma}
\newtheorem{corollary}[theorem]{Corollary}
\newtheorem{proposition}[theorem]{Proposition}
\newcommand{\blackslug}{\penalty 1000\hbox{
    \vrule height 8pt width .4pt\hskip -.4pt
    \vbox{\hrule width 8pt height .4pt\vskip -.4pt
          \vskip 8pt
      \vskip -.4pt\hrule width 8pt height .4pt}
    \hskip -3.9pt
    \vrule height 8pt width .4pt}}
\newcommand{\proofend}{\quad\blackslug}
\newenvironment{definition}{$\;$\newline \noindent {\bf Definition}$\;$}{$\;$\newline}
\def\boxit#1{\vbox{\hrule\hbox{\vrule\kern4pt
  \vbox{\kern1pt#1\kern1pt}
\kern2pt\vrule}\hrule}}
\addtolength{\baselineskip}{+0.4mm}
\def\boxit#1{\vbox{\hrule\hbox{\vrule\kern4pt
  \vbox{\kern1pt#1\kern1pt}
\kern2pt\vrule}\hrule}}

\newcommand*{\oneone}{$(1{+}1)$~\textup{EA}\xspace}
\newcommand*{\OM}{\textrm{\textup{\textsc{OneMax}}}\xspace}
\newcommand*{\nwspace}{\hspace*{.1em}}
\newcommand*{\Or}{\mathrm O}
\newcommand*{\xorg}{x_{\mathrm{org}}}
\newcommand*{\dominates}{\succcurlyeq}
\newcommand*{\sdominates}{\succ}
\newcommand*{\mutate}{\textup{mutate}}
\newcommand*{\cross}{\textup{cross}}
\newcommand*{\uar}{u.a.r.\xspace}
\newcommand*{\wrt}{w.r.t.\xspace}

\bibliographystyle{elsarticle-num}

\title{\vspace{-.4in} \bf
Time Complexity Analysis of Evolutionary Algorithms for 2-Hop (1,2)-Minimum Spanning Tree Problem\thanks{A preliminary version of this work was
   reported in the Proceedings of the 15th Workshop on Foundations of Genetic Algorithms (FOGA 2019), ACM, 2019, pp. 133-146. This work is supported by the National Natural Science Foundation of China under Grants 61802441, 61872450, 62072476, 61872048, and the Australian Research Council under Grant DP160102401.}}


\author{
 \vspace*{3mm}
 {\sc Feng Shi}$^{\mbox{\footnotesize \S}}$ \ \ \ \
 {\sc Frank Neumann}$^{\mbox{\footnotesize \textdaggerdbl}}$ \ \
 {\sc Jianxin Wang}$^{\mbox{\footnotesize \S}}$\footnote{Corresponding author, email:
 jxwang@mail.csu.edu.cn} \\ \\
 $^{\mbox{\footnotesize \S}}$School of Computer Science and Engineering \\
   Central South University \\
  \vspace*{3mm}
   Changsha 410083, P.R. China\\
   $^{\mbox{\footnotesize \textdaggerdbl}}$Department of Computer Science \\ 
   The University of Adelaide \\
    Adelaide, Australia}

\date{}
\maketitle

\vspace{-7mm}

\begin{abstract}
The Minimum Spanning Tree problem (abbr. MSTP) is a well-known combinatorial optimization problem that has been extensively studied by the researchers in the field of evolutionary computing to theoretically analyze the optimization performance of evolutionary algorithms.
Within the paper, we consider a constrained version of the problem named 2-Hop (1,2)-Minimum Spanning Tree problem (abbr. 2H-(1,2)-MSTP) in the context of evolutionary algorithms, which has been shown to be NP-hard.
Following how evolutionary algorithms are applied to solve the MSTP, we first consider the evolutionary algorithms with search points in edge-based representation adapted to the 2H-(1,2)-MSTP (including the (1+1) EA, Global Simple Evolutionary Multi-Objective Optimizer and its two variants). More specifically, we separately investigate the upper bounds on their expected time (i.e., the expected number of fitness evaluations) to obtain a $\frac{3}{2}$-approximate solution with respect to different fitness functions.
Inspired by the special structure of 2-hop spanning trees, we also consider the (1+1) EA with search points in vertex-based representation that seems not so natural for the problem and give an upper bound on its expected time to obtain a $\frac{3}{2}$-approximate solution, which is better than the above mentioned ones.
\end{abstract}

\section{Introduction}

Over the past decades, evolutionary algorithms have been extensively studied to solve the combinatorial optimization problems abstracted from real applications in various areas, including engineering, logistics, and economics. 
Although the theoretical understanding of the behavior of evolutionary algorithms (more specifically, their expected time) has achieved lots of progress, in particular, for the well-known Traveling Salesperson problem~\cite{beyer1992some,theile2009exact,sutton2014parameterized,nallaperuma2017expected,pourhassan2019theoretical,lai2020analysis}, Vertex Cover problem~\cite{oliveto2008analysis,oliveto2009analysis,friedrich2009analyses,friedrich2010approximating,yu2012approximation,kratsch2013fixed,jansen2013approximating,Pourhassan2015vc,pourhassan2016parameterized,pourhassan2017use,shi2021runtime}, Knapsack problem~\cite{kumar2005running,he2014theoretical,he2015analysis,wu2016impact,roostapour2018performance,neumann2018runtime,friedrich2018analysis}, Makespan Scheduling problem~\cite{gunia2005analysis,witt2005worst,sutton2012parameterized,neumann2015runtime,zhou2015performance,corus2019artificial,corus2021fast}, and Minimum Spanning Tree problem~\cite{neumann2007randomized,NeumannW06,kratsch2010fixed,witt2014revised,corus2013generalized,roostapour2020runtime}, etc, its development still lags far behind its success in practical applications.

Within the paper, we study the theoretical performance of evolutionary algorithms for a constrained version of the Minimum Spanning Tree problem (abbr. MSTP), aiming to get insight into their abilities to emulate the local search operations based on their populations and investigate the influence of different representations for search points on their performance. 
In the following, we start with an introduction to the background of the problem and related work in the field of evolutionary computing.
Given an edge-weighted graph $G$, the MSTP asks for a connected subgraph of $G$ that contains all vertices in $G$, without any cycle and with the minimum cost, where the cost of the subgraph is defined as the sum of the weights on its edges. 
It is well-known that the MSTP is polynomial-time solvable, using the classic Prim's algorithm~\cite{jarnik1930jistem} or Kruskal's algorithm~\cite{kruskal1956shortest}.

Neumann and Wegener~\cite{neumann2007randomized} studied the performance of the Randomized Local Search (abbr. RLS) and (1+1) EA for the MSTP. 
Taking advantage of the fitness function that penalizes the disconnectivity of the corresponding subgraph induced by the search point, the expected time of the two algorithms to obtain an optimal solution were shown to have an upper bound $O(m^2 (\log n + \log w_{max}))$ and a lower bound $\theta(m^2 \log n)$, where $m$ and $n$ denote the numbers of edges and vertices in the considered graph, respectively, and $w_{max}$ denotes the maximum weight that the edges have.
Considering the case that $w_{max}$ has a very large value, the upper bound given above was improved to $O(m^3 \log n)$ using the technique designed by Reichel and Skutella \cite{reichel2009size} to adjust the weights of the fitness function. 
Later Witt~\cite{witt2014revised} further improved the upper bound on the expected time of the (1+1) EA to $O(m^2 (\log n + \sqrt{c(G)}))$ by adaptive drift analysis with a skillful potential function, where $c(G)$ is the circumference (length of the longest cycle) of the underlying graph.
Neumann and Wegener~\cite{NeumannW06} studied the performance of the algorithm SEMO (Simple Evolutionary Multi-Objective Optimizer~\cite{laumanns2004running}) and GSEMO (Global SEMO) for the MSTP with respect to a two-objective fitness function, which consists of the number of connected components in the corresponding subgraph induced by the search point and the cost of the chosen edges.
They showed that the expected time of the two algorithms can be upper bounded by $O(mn(n + \log w_{max}))$.

Kratsch et al.~\cite{kratsch2010fixed} investigated the NP-hard problem, Maximum Leaf Spanning Tree problem (looks for a spanning tree with the maximum number of leaves), by evolutionary algorithms in the context of fixed-parameter tractability~\cite{downey2012parameterized,cygan2015parameterized}, where the number of leaves is considered as the parameter.
More specifically, they studied whether or not the considered evolutionary algorithm can obtain a feasible solution to the problem in expected FPT-runtime~\cite{kratsch2013fixed}, with respect to two mutation operators. 
Corus et al.~\cite{corus2013generalized} examined the NP-hard {\it Generalized} MSTP. 
The problem has an edge-weighted complete graph $G = (V,E,W)$ with a partition $P = \{V_1,V_2,...,V_m\}$ of the vertex set $V$ as input and asks for a subgraph of $G$ with the minimum cost that connects a vertex of each $V_i$. 
Two approaches for the bi-level optimization problem were considered: Spanned Nodes Representation and Global Structure Representation. The Spanned Nodes Representation first selects in the upper level problem a vertex (called spanned node) from each cluster $V_i$ then finds on the lower level a minimum spanning tree for these spanned nodes. 
The Global Structure Representation first constructs a complete graph $H = (V',E')$ with $m$ vertices (each one corresponds to a cluster $V_i$ in $G$) and finds in the upper level problem a spanning tree of $H$, then it selects on the lower level a spanned vertex from each cluster $V_i$.
They showed that their specific (1+1) EA working with the Spanned Nodes Representation is not a fixed-parameter evolutionary algorithm, whereas the one working with the Global Structure Representation is. 
Neumann~\cite{neumann2007expected} considered the multi-objective version of the MSTP (each edge $e$ in the input graph has a weight vector $w(e) = (w_1(e), \ldots, w_k(e))$, and $w_i(e)$ is a positive integer for all $1 \leq i \leq k$) that asks for a Pareto set that contains a minimum spanning tree with respect to each $w_i$.
He showed that a simple evolutionary algorithm can obtain a population that is a 2-approximation of the Pareto front.  
Besides the work mentioned above, there is also an active research line on evolutionary algorithms for the Bounded Diameter MSTP~\cite{raidl2003greedy,gruber2006neighbourhood,Gao2019}, where the problem looks for a spanning tree with the minimum cost such that the number of edges on any path in the tree is upper bounded by some given number.

The constrained version of the MSTP studied in this paper was abstracted from the realistic scenarios of telecommunication network construction, more specifically, connecting several sites to some central one $r$ using a spanning tree with the minimum cost.
Considering factors such as transmission delay and reliability, it is natural to constrain the number of hops (i.e., edges) in the unique path connecting any site with $r$ in the required spanning tree by an upper bound $k$, leading to the $k$-hop constraint.
Thus the realistic problem can be modeled as follows, which is well-known as {\it $k$-Hop Minimum Spanning Tree problem} (abbr. $k$H-MSTP): Given a complete graph $G= (\{r\} \cup V, E, W)$, where each vertex of $\{r\} \cup V$ corresponds to a unique site (the specific vertex $r$ corresponds to the central site), and the weight function $W$ is defined on the edge-set $E$ (the weight on each edge $[v,v']$ depends on the cost of the connection between the two sites $v$ and $v'$),  
the aim is to find a $k$-hop spanning tree with the minimum cost (i.e., a spanning tree with the minimum cost such that for any vertex $v \in V$, the unique path connecting $v$ and $r$ in it contains at most $k$ edges). 

The $k$H-MSTP has been studied in literature~\cite{deering1990multicast,deering1994architecture,kompella1993multicast,leitner2016layered,sharafeddine2017failure,das2015approximating}, and it is NP-hard as the $2$H-MSTP (i.e., $k = 2$) was shown to be NP-hard~\cite{dahl19982}.
Manyem and Stallmann showed that the $k$H-MSTP is not in APX~\cite{manyem1996some}, and Althaus et al. proposed an approximation algorithm with ratio $O(\log n)$ for it~\cite{althaus2005approximating}.
Interestingly, the $2$H-MSTP has a close relationship with the classical clustering problems.
For example, the $2$H-MSTP can be treated as a non-metric version of the well-known Uncapacitated Facility Location problem~\cite{guha1999greedy,mahdian20021}, more specifically, each vertex $v \in V$ can be chosen as a facility with open cost $W([v,r])$ or a client that should be assigned to a facility $v' \in V \setminus \{v\}$ with cost $W([v,v'])$.
Dahl~\cite{dahl19982} studied the $2$H-MSTP from a polyhedral point of view based on a directed formulation. 
Alfandari and Paschos~\cite{alfandari1999approximating} modeled a real-case network design problem facing by a French telecommunications company as the $2$H-MSTP and showed that it cannot be approximated with a ratio better than $O(\log n)$.
The famous Traveling Salesperson problem and Steiner Tree problem have been studied extensively under the $(1,2)$-setting such that each edge in the considered graph has weight 1 or 2~\cite{bern1989steiner,papadimitriou1993traveling,berman20068,angel2004approximating,baburin2009approximation} (one can consider that the weights on the edges are not well-defined, just ``small'' and ``large''). 
Thus Alfandari and Paschos~\cite{alfandari1999approximating} also investigated the 2H-MSTP under the $(1,2)$-setting and gave an approximation algorithm with ratio 5/4. Note that the NP-hardness of the $2$H-MSTP also holds even under the $(1,2)$-setting. 
To our best knowledge, there is no literature on the rigorous runtime analysis of evolutionary algorithms for the $k$H-MSTP.

In the remainder of the paper, we consider the $2$H-MSTP under the $(1,2)$-setting, called {\it 2-Hop (1,2)-Minimum Spanning Tree problem} (abbr. 2H-(1,2)-MSTP), in the context of evolutionary algorithms.
More specifically, we study the expected time (i.e., the expected number of fitness evaluations) of the considered  evolutionary algorithms to obtain an approximate solution to the problem and the corresponding approximation ratio.
Our study is to offer insight into the structural property of the problem and get hints for the analysis of evolutionary algorithms for related problems such as the $k$H-MSTP, Bounded Diameter MSTP, Cluster Median problem, and Facility Location problem.

\begin{table}[]
\begin{center}
\renewcommand{\arraystretch}{1.2}
\begin{tabular}{c|c|c|c|c}
\toprule
\multicolumn{2}{c|}  {\textbf{ }}       & \multicolumn{2}{c|}{Edge-based Representation}   & Vertex-based Representation   \\ \hline
\multicolumn{2}{c|}{\textbf{Ratio}}     & 2 (worst case)                & 3/2 & 3/2   \\ \hline
\multicolumn{2}{c|}{\textbf{(1+1) EA}}  &  $O(m \log n)$    &  $O(m^6 n)$   &  $O(n^4)$    \\ 
\multicolumn{2}{c|}{\textbf{GSEMO}}      &  $O(mn \log n)$  &  $O(m^6 n^2)$  & /   \\ 
\multicolumn{2}{c|}{\textbf{GSEMO-1}}    &  $O(mn)$          &  $O(m^6 n)$   & /    \\ 
\multicolumn{2}{c|}{\textbf{GSEMO-2}}    &  $O(m \log n)$    &  $O(m^4 n)$   & /  \\ 
\bottomrule
\end{tabular}
\end{center}
\caption{Overview of the upper bounds on the expected time of the four algorithms with search points in edge-based representation (namely, the \oneone, Global Simple Evolutionary Multi-Objective Optimizer (GSEMO) and its two variants (GSEMO-1 and GSEMO-2)) and the \oneone with search points in vertex-based representation to get an approximate solution to the 2H-(1,2)-MSTP with ratio 2 or 3/2. 
Variables $m$ and $n+1$ denote the numbers of edges and vertices in the considered graph, respectively, and $m = \Theta(n^2)$.}
\label{table:overviewResults}
\vspace*{-2mm}
\end{table}

Literature~\cite{raidl2000efficient,ashlock2012representation,corus2013generalized,hu2015new} shows that the choice of representations for search points has an enormous impact on the performance of evolutionary algorithms.
Thus, apart from considering the edge-based representation for search points in this paper (i.e., the algorithms look for the edges that are in the optimal solution to the 2H-(1,2)-MSTP), we also consider the vertex-based representation for search points that seems not so natural for the problem (i.e., the algorithms look for the neighbors of $r$ in the optimal solution to the 2H-(1,2)-MSTP), based on the fact that the optimal solution is easy to construct if the neighbors of $r$ in it are known.

Since a search point (or solution) in edge-based representation may not be {\it feasible} (a solution is {\it feasible} if it corresponds to a 2-hop spanning tree), we first analyze the performance of the classic (1+1) EA (with search points in edge-based representation) to get a feasible solution to the 2H-(1,2)-MSTP with respect to a designed fitness function. 
Assume the considered graph $G= (\{r\} \cup V, E, W)$ contains $n+1$ vertices and $m$ edges. 
Due to the elitist selection mechanism, if the (1+1) EA has obtained a feasible solution, then it needs to ``swap'' an edge in the spanning tree with an edge not in it (i.e., takes expected time $O(m^2)$) to further improve the maintained solution without introducing any cycle or causing any disconnectivity. 
Thus we also study a multi-objective evolutionary algorithm, called Global Simple Evolutionary Multi-Objective Optimizer (abbr. GSEMO, with search points in edge-based representation), aiming to avoid the ``swap'' operation by maintaining a population. 
The GSEMO keeps a solution with $i$ edges for each $i \in [0,n]$ (i.e., its population size is at most $n+1$) and can construct a 2-hop spanning tree in the greedy way (as the way of Prim's algorithm and Kruskal's algorithm).  
However, the large population of the GSEMO may slow down its optimization process, hence we present two variants of the GSEMO, named GSEMO-1 and GSEMO-2, where each of them maintains a population with at most two solutions.
We show that the GSEMO-1 and GSEMO-2 can efficiently emulate the ``swap'' operation and the local search operation with more than two edges, respectively.  
Afterwards, using the local search technique, we separately analyze the upper bounds on the expected time of the four algorithms mentioned above to get an approximate solution with ratio 3/2. 
A summary of the obtained results is given in Table~\ref{table:overviewResults}.

With respect to the vertex-based representation, we only consider the (1+1) EA (with search points in vertex-based representation) and show that its performance to get an approximate solution with ratio 3/2 is much better than that of the above mentioned algorithms. 
There are two reasons for its better performance: 
(1) the edges are assumed to be connected in a way with the minimum cost based on the neighbors of $r$ specified by the considered solution, i.e., all solutions in vertex-based representation are feasible; 
(2) an algorithm with search points in vertex-based representation is more efficient than one with search points in edge-based representation to emulate a local search operation.
For example, consider a feasible solution $x$ with two vertices $v$ and $v'$ that are neighbor and not neighbor to $r$, respectively, and a local search operation on $x$ that makes $v$ and $v'$ not neighbor and neighbor to $r$, respectively, in the resulting feasible solution. 
Then obviously, the (1+1) EA with search points in vertex-based representation and the one with search points in edge-based representation take expected time $O(n^2)$ and $O(m^{2c})$, respectively, to emulate the operation, where $c \ge 2$ is the number of edges incident to $v$ or $v'$ in the graph specified by $x$ (as the (1+1) EA with search points in edge-based representation considers not only the removal but also the addition of edges to make the resulting solution feasible).

This paper extends and refines the conference version~\cite{DBLP:conf/foga/00030W19}.
Firstly, the discussion for the vertex-based representation is newly added. 
Secondly, for the four algorithms with search points in edge-based representation (including the (1+1) EA, GSEMO and its two variants), this paper replaces the previously complicated fitness functions with new ones (Section 3) and gives a completely new discussion for their performance to get a feasible solution (Section 4).
Finally, more details of the algorithms and illustrations are included.  

The rest of the paper is organized as follows.
Section 2 introduces related definitions. Section 3 presents the four considered algorithms with search points in edge-based representation, namely, \oneone, GSEMO and its two variants, and Section 4 analyzes their performance to obtain a feasible solution. 
The in-depth analysis for their performance to get an approximate solution to the 2H-(1,2)-MSTP with ratio 3/2 is given in Section 5.
Section 6 considers the \oneone with search points in vertex-based representation and its performance to get an approximate solution to the 2H-(1,2)-MSTP with ratio 3/2.
Section 7 is used to conclude this work.

\section{Preliminaries}
\label{sec:prelims}
A graph is {\it complete} if there is an edge between any two vertices in the graph.
Consider a complete edge-weighted graph $G=(\{r\} \cup V,E,W)$, where $r$ is a specific vertex that is called the {\it root} of $G$ in the remaining context, $V = \{v_1,\ldots,v_n\}$, $E= \{e_1,\ldots,e_m\}$, and $W : E \rightarrow \{1,2\}$ (i.e., $G$ has $n+1$ vertices including $r$ and $m$ edges). 
For any vertex $v \in \{r\} \cup V$, denote by $N_i(v)$ ($i \in \{1,2\}$) the set containing all the vertices $v' \in (\{r\} \cup V) \setminus \{v\}$ with $W([v,v']) = i$.
For any edge-subset $E' \subset E$, denote by $G[E']$ the graph obtained by removing all edges in $E \setminus E'$ from $G$ (i.e., $G[E']$ and $G$ have the same vertex-set $\{r\} \cup V$).   

A {\it spanning tree} $T$ of $G = (\{r\} \cup V, E, W)$ is a subgraph of $G$ that connects all vertices in $\{r\} \cup V$ and has no cycle. 
The {\it weight} of $T$ is the sum of the weights on its edges.
A spanning tree of $G$ is the {\it minimum} if it has the minimum weight among all spanning trees of $G$. 
The {\it Minimum Spanning Tree problem} (abbr. MSTP) on $G$ looks for a subset $E^*$ of $E$ such that $G[E^*]$ is a minimum spanning tree of $G$.
As each edge in the considered graph $G$ has weight 1 or 2, the Minimum Spanning Tree problem on $G$ is also called the (1,2)-Minimum Spanning Tree problem (abbr. (1,2)-MSTP) on $G$ in the context.

In the paper a constrained variant of the (1,2)-MSTP on $G = (\{r\} \cup V, E, W)$ is studied, named {\it 2-Hop (1,2)-Minimum Spanning Tree problem} (abbr. 2H-(1,2)-MSTP), which looks for a minimum spanning tree $T^*$ of $G$ satisfying the {\it 2-hop constraint}. 
Recall that the 2-hop constraint requires that for any vertex $v \in V$, the unique path connecting $v$ and $r$ in $T^*$ contains at most two edges.

As mentioned in the previous section, two representations for search points are considered: Edge-based representation and vertex-based representation. 
The search space corresponding to the edge-based representation consists of all bit-strings with fixed length $m$. 
That is, for a search point $x = x_1 \ldots x_m$ (in edge-based representation), the edge $e_i$ ($1 \leq i \leq m$) is chosen if and only if $x_i = 1$. 
Denote by $E(x)$ the subset of $E$ containing all edges chosen by $x$. 
For simplicity of notation, let $G(x)$ be the same graph as $G[E(x)]$. 
Denote the {\it cost} of $x$ by
$$c(x) = \sum_{i = 1}^{m} W(e_i) \cdot x_i.$$

The search space corresponding to the vertex-based representation consists of all bit-strings with fixed length $n$.
That is, for a search point $x = x_1 \ldots x_n$ (in vertex-based representation), the vertex $v_i$ ($1 \leq i \leq n$) is chosen if and only if $x_i = 1$. 
Denote by $V(x)$ the subset of $V$ containing all vertices chosen by $x$. 
Now based on $V(x)$, an unambiguous way to construct a 2-hop spanning tree $G(x)$ with the minimum cost is given as follows: firstly remove all edges of $E$ from $G$ except the edges between $r$ and the vertices of $V(x)$; secondly for each vertex $v \in V \setminus V(x)$, if there is a vertex $v' \in V(x)$ with $W([v,v']) = 1$, then connect $v$ to $v'$ with an edge, otherwise, connect $v$ to an arbitrarily vertex $v' \in V(x)$ with an edge.
Denote the {\it cost} of $x$ (i.e., the sum of weights on the edges in $G(x)$) by
$$c(x) = \sum_{e \in G(x)} W(e).$$

Note that for any search point $x$ in vertex-based representation, $G(x)$ is always a 2-hop spanning tree of $G$, i.e., $x$ is always a {\it feasible} solution. 
However, for a search point $x$ in edge-based representation, $G(x)$ may not be a 2-hop spanning tree of $G$, i.e., $x$ may be an {\it infeasible} solution.
Thus we have the following notations for the search point $x$ in edge-based representation. 

Denote by $|x|_1$ the number of 1-bits in $x$ (i.e., the Hamming weight of $x$), $C_r(x)$ the connected component in $G(x)$ that contains the root $r$, and $N_{\textup{cc}}(x)$ the number of connected components in $G(x)$. 
Given two vertices $v_1$ and $v_2$ in $G(x)$, the {\it distance} between them, denoted by $d_{G(x)}(v_1,v_2)$, is defined as the number of edges on the shortest path connecting $v_1$ and $v_2$ in $G(x)$ if they are in the same connected component of $G(x)$; otherwise, $n+1$ (as $G$ contains $n+1$ vertices, the distance between $v_1$ and $v_2$ ranges from $1$ to $n$ if $v_1$ is in the same connected component with $v_2$).
Denote by $N_{d > i}(x)$ the number of vertices $v \in V$ with $d_{G(x)}(v,r) > i$, where $1 \le i \le n$ is an integer (i.e., $N_{d > n}(x)$ is the number of vertices that are not in the same connected component $C_r(x)$ with $r$ in $G(x)$). 
Denote by $N_{n \ge d > 2}(x)$ the number of vertices where they are in the same connected component $C_r(x)$ with $r$ in $G(x)$, but their distances to $r$ are greater than 2. 

Using the notations given above, the 2H-(1,2)-MSTP can be reformulated as looking for a search point $x$ in edge-based representation with the minimum cost such that $|x|_1 = n$ and $d_{G(x)}(v,r) \leq 2$ for any vertex $v \in V$.

\section{Algorithms with Search Points in Edge-based Representation}

In the section, we present four evolutionary algorithms with search points in edge-based representation, namely, $(1{+}1)$ EA, Global Simple Evolutionary Multi-Objective Optimizer (abbr. GSEMO) and its two variants (GSEMO-1 and GSEMO-2).

\subsection{\oneone}
\label{subsec:prelims_algo}

The $(1{+}1)$ EA (given in Algorithm~\ref{alg:1+1}) starts with an arbitrary solution.
In each iteration, the algorithm generates an offspring $y$ based on the maintained solution $x$ using standard bit mutation and accepts $y$ if it is at least as good as $x$. 
The scalar-valued fitness function $f_{1+1}(x)$ exploited by the (1+1) EA is given below, which not only considers the cost of the solution $x$ (term $c(x)$) but also penalizes the vertices whose distances to $r$ are greater than 2 (term $N_{d > 2}(x)$) and the excess Hamming weight (term $\max\{|x|_1 - n,0\}$) with penalization coefficients $2m^2$ and $m^2$, respectively. 
$$f_{1+1}(x) = c(x) + m^2 \cdot \left(2 N_{d > 2}(x) + \max\{|x|_1 - n,0\}\right)$$

\begin{algorithm2e}
\caption{\oneone}
\label{alg:1+1}
\SetAlgoSkip{tinyskip}
  choose $x \in \{0,1\}$ uniformly at random\;
  determine $f_{1+1}(x)$\;
  \While{stopping criterion not met}{
    $y$ $\gets$ flip each bit of $x$ independently with probability $1/m$\;
    determine $f_{1+1}(y)$\;
    \If {$f_{1+1}(y) \leq f_{1+1}(x)$}{$x \gets y$\;}
  }
\end{algorithm2e}

Recall that the graph $G=(\{r\} \cup V,E,W)$ considered in the paper is complete (i.e., $m = \theta(n^2)$), and $c(x) \leq 2m$, thus the two penalization coefficients $2m^2$ and $m^2$ are large enough.
The extra factor $2$ of the penalization coefficient of the term $N_{d > 2}(x)$ compared to that of the term $\max\{|x|_1 - n,0\}$ helps the algorithm to make a trade-off. 

Observe that although the fitness function $f_{1+1}(x)$ guides the (1+1) EA to find a feasible solution, it has few benefits in improving a feasible solution.
For example, assume that the (1+1) EA has obtained a feasible solution $x$. As any infeasible solution cannot be accepted by the algorithm ever again, the (1+1) EA needs to apply the ``swap'' operation to construct a new feasible solution, where the ``swap'' operation replaces an edge of $E(x)$ with one of $E \setminus E(x)$ (i.e., flip a 1-bit and a 0-bit in $x$). 
Each execution of the ``swap'' operation takes expected time $O(m^{2})$ and cannot be decomposed (i.e., the edge-removal operation and the edge-addition operation should be accomplished at the same time; otherwise, an infeasible solution is constructed). 
Thus we also consider several multi-objective evolutionary algorithms and try to use their populations to avoid the expensive runtime caused by the ``swap'' operation.

\subsection{Multi-Objective Evolutionary Algorithms}

The Global Simple Evolutionary Multi-Objective Optimizer (abbr. GSEMO, given in Algorithm~\ref{alg:GSEMO}) uses a vector-valued fitness function
\begin{equation*}
    f_{\textup{M}}(x) = \left[|x|_1, f(x)\right],
\end{equation*}
where $f(x) = c(x) + m^2 \cdot N_{d > 2}(x)$ (as the first element of the vector considers the Hamming weight of $x$, the second one $f(x)$ can ignore the term $\max\{|x|_1 - n,0\}$ that is considered in $f_{1+1}(x)$). 
Any optimal solution (in edge-based representation) to the 2H-(1,2)-MSTP has exactly Hamming weight $n$, thus the GSEMO can only keep the solutions with Hamming weights ranging from 0 to $n$ and works in an incremental way to find a feasible solution (as the way of Prim's algorithm and Kruskal's algorithm).
For the solutions with Hamming weights over $n$ (as the initial population of the algorithm contains an arbitrary solution), they are defined to be {\it dominated} by the solutions with Hamming weights in $[0,n]$, where the definition of the {\it dominance} in the context of the GSEMO is given as follows.
Given two solutions $y$ and $z$, if \emph{at most one} of the values $|y|_1$ and $|z|_1$ is in $[0, n]$, then we set 
\begin{equation*}
\label{eq:dominance_GSEMO-1}
  y \dominates_{\mathrm{GSEMO}} z \ \Leftrightarrow \ (|y|_1 < |z|_1) \,\vee\, (|y|_1 = |z|_1 \wedge f(y) \le f(z)); 
\end{equation*}
\noindent otherwise (i.e., both $|y|_1$ and $|z|_1$ are in $[0, n]$),
\begin{equation*}
\label{eq:dominance_GSEMO-1_2}
  y \dominates_{\mathrm{GSEMO}} z \ \Leftrightarrow\ |y|_1 = |z|_1 \,\wedge\, f(y) \le f(z),
\end{equation*}
where $y \dominates_{\mathrm{GSEMO}} z$ says that $y$ {\it dominates} $z$ with respect to $\dominates_{\mathrm{GSEMO}}$. Solution $y$ {\it strongly dominates} $z$ if $y \dominates_{\mathrm{GSEMO}} z$ but $f_{\textup{M}}(y) \neq f_{\textup{M}}(z)$, written $y \sdominates_\textup{GSEMO} z$.  

The GSEMO starts with a population $S$ that contains an arbitrary solution.
In each iteration, the GSEMO picks an individual $x$ randomly from $S$ and generates an offspring $y$ based on $x$ using standard bit mutation. 
If $y$ is not strongly dominated by a solution in $S$ with respect to $\sdominates_{\mathrm{GSEMO}}$, then all solutions dominated by $y$ with respect to $\dominates_{\mathrm{GSEMO}}$ in $S$ are discarded, and $y$ is included into $S$. 
By the {\it dominance} given above with respect to $\dominates_{\mathrm{GSEMO}}$, if $S$ has a solution $x$ with Hamming weight in $[0,n]$, then $S$ has neither a solution with the same Hamming weight $|x|_1$ nor a solution with Hamming weight over $n$; otherwise, the size of $S$ is exactly 1 (as any two solutions with Hamming weights over $n$ are comparable with respect to $\dominates_{\mathrm{GSEMO}}$).
Thus the size of $S$ is upper bounded by $n+1$. 

\begin{algorithm2e}
\caption{GSEMO}
\label{alg:GSEMO}
\SetAlgoSkip{tinyskip}
    choose $x \in \{0,1\}^m$ uniformly at random\;
    determine $f_{\textup{M}}(x)$\;
    $S \leftarrow \{x\}$\;
    \While{stopping criterion not met}{
        choose $x \in S$ uniformly at random\;
        $y$ $\gets$ flip each bit of $x$ independently with probability $1/m$\;
        determine $f_{\textup{M}}(y)$\;
        \If {$\nexists \nwspace w \in S \colon w \sdominates_{\textup{GSEMO}} y$}{
            $S \gets (S \setminus \{z \in S \mid y \dominates_{\textup{GSEMO}} z \}) \cup \{y\}$\;
        }
   }
\end{algorithm2e}

Unfortunately, the large size $n+1$ of the population maintained by the GSEMO may slow down its optimization process, hence we present a variant of it, named GSEMO-1 (given in Algorithm~\ref{alg:GSEMO-1}).
The GSEMO-1 uses the same fitness function $f_{\textup{M}}$ as GSEMO but differs at the notion of dominance between solutions, written $\dominates_{\mathrm{GSEMO-1}}$, whose definition is inspired by that of the dominance $\dominates_{\mathrm{GSEMO-S}}$ given in~\cite{shi2019reoptimization}.
Given two solutions $y$ and $z$, if \emph{at most one} of the values $|y|_1$ and $|z|_1$ is in $[n, n+1]$, then we set
\begin{equation*}
\label{eq:dominance_GSEMO-1}
  y \dominates_{\mathrm{GSEMO-1}} z \ \Leftrightarrow \ (||y|_1 - n| < ||z|_1 - n|) \,\vee\, (|y|_1 = |z|_1 \wedge f(y) \le f(z));
\end{equation*}

\noindent
otherwise,
\begin{equation*}
\label{eq:dominance_GSEMO-1_2}
  y \dominates_{\mathrm{GSEMO-1}} z \ \Leftrightarrow\ |y|_1 = |z|_1 \,\wedge\, f(y) \le f(z),
\end{equation*}
where $y \dominates_{\mathrm{GSEMO-1}} z$ says that $y$ {\it dominates} $z$ with respect to $\dominates_{\mathrm{GSEMO-1}}$. Solution $y$ {\it strongly dominates} $z$ if $y \dominates_{\mathrm{GSEMO-1}} z$ but $f_{\textup{M}}(y) \neq f_{\textup{M}}(z)$, written $y \sdominates_{\mathrm{GSEMO-1}} z$. 

Consequently, two bit strings $y$ and $z$ are incomparable if and only if both $|y|_1$ and $|z|_1$ are in $[n, n+1]$ and $|y|_1 \neq |z|_1$, implying that the size of the population maintained by the GSEMO-1 can be upper bounded by 2.
The purpose to keep such a population that can maintain two solutions with Hamming weights $n$ and $n+1$ respectively at the same time, is that the solution with Hamming weight $n+1$ can be an intermediate to accelerate the optimization process of the solution with Hamming weight $n$.
More specifically, consider a ``swap'' operation on the feasible solution $x$ maintained by the GSEMO-1 that replaces an edge $e \in E(x)$ with an edge in $e' \in E \setminus E(x)$ such that the resulting solution has a better fitness than $x$.
Its execution can be split into two separate steps: The first step is including the edge $e'$, which results in an infeasible solution with Hamming weight $n+1$ (assume that the infeasible solution can be accepted by the algorithm); the second step is removing the edge $e$ from the infeasible solution to construct a feasible solution. 
As both the first and second step can be accomplished in runtime $O(m)$, the ``swap'' operation can be emulated within runtime $O(m)$, not $O(m^2)$.

\begin{algorithm2e}
\caption{GSEMO-1}
\label{alg:GSEMO-1}
\SetAlgoSkip{tinyskip}
    choose $x \in \{0,1\}^m$ uniformly at random\;
    determine $f_{\textup{M}}(x)$\;
    $S \leftarrow \{x\}$\;
    \While{stopping criterion not met}{
        choose $x \in S$ uniformly at random\;
        $y$ $\gets$ flip each bit of $x$ independently with probability $1/m$\;
        determine $f_{\textup{M}}(y)$\;
        \If {$\nexists \nwspace w \in S \colon w \sdominates_{\textup{GSEMO-1}} y$}{
            $S \gets (S \setminus \{z \in S \mid y \dominates_{\textup{GSEMO-1}} z \}) \cup \{y\}$\;
        }
   }
\end{algorithm2e}

Inspired by the idea of the GSEMO-1 that maintains an extra infeasible solution with Hamming weight $n+1$, we design an algorithm that maintains an extra infeasible solution such that replacing exactly one of its edges with a new edge can generate a feasible solution, to further accelerate the optimization process. Thus we have the following notations.

Given a solution $x$, denote by $V_d(x)$ the subset of $V$ with the minimum size such that adding the edges between $r$ and the vertices of $V_d(x)$ into $G(x)$ obtains a solution $x'$ with $N_{n \geq d > 2}(x') = 0$.
Observe that if $N_{\textup{cc}}(x) = |V_d(x)| = 1$, then a solution $x'$ with $N_{\textup{cc}}(x') =  1$ and $|V_d(x')| = 0$ (i.e., $N_{d > 2}(x') = 0$) can be obtained by adding the edge $[v,r]$ into $G(x)$ with $v \in V_d(x)$.
Note that $G(x')$ may contain cycles, but the edges whose removal from $G(x')$ obtains a feasible solution is easy to find. 
Thus we can use the solutions $x$ with $N_{\textup{cc}}(x) = |V_d(x)| = 1$ as intermediates to promote the optimization process of the maintained solutions. The second variant of the GSEMO, named GSEMO-2, is given in Algorithm~\ref{alg:GSEMO-2}, using the following vector-valued fitness function 
$$f_{\textup{M2}}(x) = \left[f_{\textup{M2}}^1(x), f_{\textup{M2}}^2(x) \right],$$
where $f_{\textup{M2}}^1(x) = |V_d(x)| + m^2 \cdot (N_{\textup{cc}}(x) - 1) \ \textup{and} \ f_{\textup{M2}}^2(x) = c(x) + m^2 \cdot \max\{|x|_1 - n, 0\}$.

Given two solutions $y$ and $z$, if \emph{at most one} of $f_{\textup{M2}}^1(y)$ and $f_{\textup{M2}}^1(z)$ is in $[0, 1]$, then we set
\begin{equation*}
\label{eq:dominance_GSEMO-1}
  y \dominates_{\mathrm{GSEMO-2}} z \ \Leftrightarrow \ (f_{\textup{M2}}^1(y) < f_{\textup{M2}}^1(z)) \,\vee\, (f_{\textup{M2}}^1(y) = f_{\textup{M2}}^1(z) \,\wedge\, f_{\textup{M2}}^2(y) \le f_{\textup{M2}}^2(z));
\end{equation*}
otherwise, \begin{equation*}
\label{eq:dominance_GSEMO-1_2}
  y \dominates_{\mathrm{GSEMO-2}} z \ \Leftrightarrow\ f_{\textup{M2}}^1(y) = f_{\textup{M2}}^1(z) \,\wedge\, f_{\textup{M2}}^2(y) \le f_{\textup{M2}}^2(z),
\end{equation*}
\noindent where $y \dominates_{\mathrm{GSEMO-2}} z$ says that $y$ {\it dominates} $z$ with respect $\dominates_{\mathrm{GSEMO-2}}$.
Solution $y$ {\it strongly dominates} $z$ if $y \dominates_{\mathrm{GSEMO-2}} z$ but $f_{\textup{M2}}(y) \neq f_{\textup{M2}}(z)$, written $y \sdominates_\textup{GSEMO-2} z$.
Consequently, two bit strings $y$ and $z$ are incomparable if and only if both $f_{\textup{M2}}^1(y)$ and $f_{\textup{M2}}^1(z)$ are in $[0, 1]$ and $f_{\textup{M2}}^1(y) \neq f_{\textup{M2}}^1(z)$, implying that the size of the population maintained by the GSEMO-2 can be upper bounded by 2.

Regarding the way to decide whether $0 \le f_{\textup{M2}}^1(x) \le 1$, we first have the observation that $f_{\textup{M2}}^1(x) =0$ if and only if $N_{\textup{cc}}(x) = 1$ and $N_{n \ge d > 2}(x) = 0$, and $f_{\textup{M2}}^1(x) \ge m^2$ if and only if $N_{\textup{cc}}(x) > 1$.
Thus the remaining issue is to decide whether $f_{\textup{M2}}^1(x) = 1$ under the assumption that $N_{\textup{cc}}(x) = 1$ and $f_{\textup{M2}}^1(x) \neq 0$.
Unfortunately, it is tedious to list all possible structures of $G(x)$ satisfying $f_{\textup{M2}}^1(x) = 1$, so we give a simple way to decide whether $f_{\textup{M2}}^1(x) = 1$: For each vertex $v \in V$ that is not a neighbor of $r$ in $G(x)$, adding the edge between $v$ and $r$ into $G(x)$ and deciding whether $N_{n \ge d > 2}(x') = 0$ for the resulting solution $x'$.
If there is such a vertex $v$, then obviously $f_{\textup{M2}}^1(x) = 1$; otherwise, $f_{\textup{M2}}^1(x) > 1$.

\begin{algorithm2e}
\caption{GSEMO-2}
\label{alg:GSEMO-2}
\SetAlgoSkip{tinyskip}
    choose $x \in \{0,1\}^m$ uniformly at random\;
    determine $f_{\textup{M2}}(x)$\;
    $S \leftarrow \{x\}$\;
    \While{stopping criterion not met}{
        choose $x \in S$ uniformly at random\;
        $y$ $\gets$ flip each bit of $x$ independently with probability $1/m$\;
        determine $f_{\textup{M2}}(y)$\;
        \If {$\nexists \nwspace w \in S \colon w \sdominates_{\textup{GSEMO-2}} y$}{
            $S \gets (S \setminus \{z \in S \mid y \dominates_{\textup{GSEMO-2}} z \}) \cup \{y\}$\;
        }
   }
\end{algorithm2e}

\section{Performance of the Four Algorithms with Search Points in Edge-based Representation for Feasible Solutions}

As $G = (\{r\} \cup V,E,W)$ has $n+1$ vertices, and each edge in $E$ has weight 1 or 2, the weight of a spanning tree in $G$ ranges from $n$ to $2n$. Thus we have the observation given below. 

\medskip
\noindent{\bf Observation.} Any feasible solution to the 2H-(1,2)-MSTP on $G$ has a worst-case approximation ratio 2. 
\medskip

Given a solution $x$ with $N_{\textup{cc}}(x) + |x|_1 > n +1$ (i.e., $G(x)$ has at least one cycle), the following lemma indicates that $G(x)$ contains $N_{\textup{cc}}(x) + |x|_1 - n -1$ edges such that removing any of them obtains a better solution. 

\begin{lemma}
\label{lem:update solution with weight greater than n}
Given a solution $x$ with $N_{\textup{cc}}(x) + |x|_1 > n +1$, it contains $N_{\textup{cc}}(x) + |x|_1 - n -1$ many 1-bits, each of whose flip results in a new solution $x'$ with $c(x') < c(x)$, $N_{\textup{cc}}(x') = N_{\textup{cc}}(x)$, and $N_{d>2}(x') = N_{d>2}(x)$.
\begin{proof}
Let $C$ be an arbitrary cycle in $G(x)$ (note that $C$ contains at least three edges as the considered graph $G$ is a simple graph). 
If $C$ is not in the connected component $C_r(x)$ that contains $r$, then for the solution $x'$ obtained by a mutation that flips exactly one of the 1-bits in $x$ corresponding to the edges in $C$, it satisfies the claimed conditions. 
The discussion for the situation that $C$ is in the connected component $C_r(x)$ is divided into two cases.  

Case (1). All vertices in $C$ have distances not greater than 1 to $r$ in $G(x)$. 
That is, for any vertex $v \in C \setminus \{r\}$, there always exists an edge between $v$ and $r$ in $G(x)$. 
As $C$ contains at least three edges, it contains an edge $e = [v_1,v_2]$, where neither $v_1$ nor $v_2$ is $r$. 
Additionally, we can derive that for any vertex $v \in C_r(x)$, the shortest path connecting $v$ and $r$ in $G(x)$ cannot contain the edge $e$. 
Thus the removal of $e$ would not change the distances of the vertices in $V$ to $r$.
For the solution obtained by the mutation that flips the 1-bit corresponding to the edge $e$ in $x$ and nothing else, it satisfies the claimed conditions. 

Case (2). There exists a vertex $v \in C$ whose distance to $r$ is not less than 2.
Let $P$ be a shortest path connecting $v$ and $r$ in $G(x)$, and $v_1$ be a neighbor of $v$ in $G(x)$ that is in $C$, but not in $P$. 
Note that $v_1$ cannot be the root~$r$. 
Denote by $x'$ the solution obtained by the mutation flipping the 1-bit corresponding to $[v,v_1]$ in $x$ and nothing else.
In the following discussion, we show that $N_{d>2}(x) = N_{d>2}(x')$. 
Firstly, as $G(x')$ is a subgraph of $G(x)$, we have that 
\begin{eqnarray}
\label{eqn:4.1-1}
N_{d>2}(x) \leq N_{d>2}(x').
\end{eqnarray}

Now we assume that there exists a vertex $v' \in V \setminus \{v\}$ with $d_{G(x)}(v',r) \leq 2$, but $d_{G(x')}(v',r) > 2$. 
That is, any shortest path connecting $v'$ and $r$ in $G(x)$ always goes through the edge $[v,v_1]$, implying that $d_{G(x)}(v',r) \ge d_{G(x)}(v,r)+1 \ge 3$, a contradiction to the assumption. 
In other words, for any vertex $v' \in V \setminus \{v\}$, if $d_{G(x)}(v',r) \leq 2$ then $d_{G(x')}(v',r) \leq 2$. Thus
$N_{2 \geq d > 0}(x) \leq N_{2 \geq d > 0}(x')$ and  
\begin{eqnarray}
\label{eqn:4.1-2}
N_{d>2}(x) \geq N_{d>2}(x').
\end{eqnarray}

By Inequalities~(\ref{eqn:4.1-1}) and~(\ref{eqn:4.1-2}), we have $N_{d>2}(x) = N_{d>2}(x')$.
Then combining the equality with the obvious fact that $c(x) > c(x')$ and $N_{\textup{cc}}(x) = N_{\textup{cc}}(x')$ shows that $x'$ satisfies the claimed conditions.

Now we show the existence of $N_{\textup{cc}}(x) + |x|_1 - n -1$ many such 1-bits in $x$. Initialize an edge-set $E_s = \O$.
Let $C$ be an arbitrary cycle in $G(x) \setminus E_s$, and $e$ be the edge in $C$ found by the way mentioned above. Include the edge $e$ into $E_s$.
Repeat the above process until $G(x) \setminus E_s$ has no cycle. 
Then it is easy to see that $E_s$ contains $N_{\textup{cc}}(x) + |x|_1 - n -1$ many edges, and the $N_{\textup{cc}}(x) + |x|_1 - n -1$ many 1-bits in $x$ corresponding to the edges in $E_s$ are obtained.
\end{proof}
\end{lemma}

\subsection{$(1{+}1)$ EA with Search Points in Edge-based Representation}

In the subsection, we study the performance of the \oneone with search points in edge-based representation to get a feasible solution to the 2H-(1,2)-MSTP on $G = (\{r\} \cup V,E,W)$.

\begin{theorem}
\label{thm:performance (1+1) EA 2}
The $(1{+}1)$ EA with search points in edge-based representation takes expected time $O(m \log n)$ to obtain a feasible solution to the 2H-(1,2)-MSTP on $G = (\{r\} \cup V,E,W)$.

\begin{proof}
Since the algorithm starts with an arbitrary solution $x$, the case of $f_{1+1}(x) \geq m^2$ (i.e., $2 N_{d>2}(x) + \max \{|x|_1 - n, 0\} > 0$) should be considered.
Let $P(x) = 2 N_{d>2}(x) + \max \{|x|_1 - n, 0\}$ be the potential of $x$, using which we analyze the expected time of the algorithm to obtain a solution $x^{\dag}$ with $P(x^{\dag}) = 0$ (note that $x^{\dag}$ is a feasible solution).
The following discussion is divided into three cases based on the value of $|x|_1$.

Case (I). $|x|_1 < n$ (i.e., $\max \{|x|_1 - n, 0\} = 0$).
For any vertex $v$ with $d_{G(x)}(v,r) >2$, flipping the 0-bit corresponding to the edge between $v$ and $r$ in $x$ generates a new solution $x'$ with $|x'|_1 = |x|_1 +1$ and $N_{d>2}(x') \le N_{d>2}(x)-1$.
Meanwhile, since $|x|_1 < n$, $P(x') \le P(x)- 2$ and $f_{1+1}(x') \le f_{1+1}(x) + 2 -2m^2$, implying that the solution $x'$ can be accepted by the algorithm. 
Considering the $N_{d>2}(x) = P(x)/2$ many 0-bits in $x$ that correspond to the edges between $r$ and the vertices $v$ with $d_{G(x)}(v,r) >2$, standard bit mutation selects exactly one of them and nothing else with probability $\Omega(\frac{P(x)}{e \cdot m})$.
Thus combining the probability with the potential decrement mentioned above gives a drift in the potential of $\Omega(\frac{P(x)}{e \cdot m})$.

Case (II). $|x|_1 = n$ (i.e., $\max \{|x|_1 - n, 0\} = 0$). If $N_{d>2}(x) =0$ then the algorithm obtains a feasible solution. Thus in the following discussion, we assume that $N_{d>2}(x) > 0$.
For any vertex $v$ with $d_{G(x)}(v,r) >2$, flipping the 0-bit corresponding to the edge between $v$ and $r$ in $x$ generates a new solution $x'$ with $N_{d>2}(x') \le N_{d>2}(x)-1$. 
However, the flip causes $\max \{|x'|_1 - n, 0\} = 1$, thus  
 $P(x') \le P(x)-1$ and $f_{1+1}(x') \le f_{1+1}(x) + 2 - m^2$, implying that the solution $x'$ can be accepted by the algorithm.
Using the drift analysis similar to that given for Case (I), we can derive that the drift in the potential is $\Omega(\frac{P(x)}{e \cdot m})$ as well.

Case (III). $|x|_1 > n$ (i.e., $\max \{|x|_1 - n, 0\} > 0$). 
Lemma~\ref{lem:update solution with weight greater than n} shows that $x$ has $N_{\textup{cc}}(x) + |x|_1 - n -1 \ge |x_1| - n$ many 1-bits, each of whose flip results in a solution $x'$ with $|x'|_1 = |x|_1 -1$, $c(x') < c(x)$ and $N_{d>2}(x') = N_{d>2}(x)$. Thus $P(x') \le P(x)-1$ and $f_{1+1}(x') < f_{1+1}(x) - m^2$, implying that the solution $x'$ can be accepted by the algorithm. 
Additionally, the analysis similar to that given for Case (II) shows that $x$ has $N_{d>2}(x)$ 0-bits, each of whose flip results in a solution $x'$ with $|x'|_1 = |x|_1 +1$, $P(x') \le P(x)-1$ and $f_{1+1}(x') \le f_{1+1}(x) + 2 - m^2$. Thus the solution $x'$ can be accepted by the algorithm as well. 
Considering all the $|x_1| - n$ 1-bits and $N_{d>2}(x)$ 0-bits in $x$ (i.e., at least $P(x)/2$ bits of $x$ in total), standard bit mutation selects exactly one of them and nothing else with probability $\Omega(\frac{P(x)}{e \cdot m})$.
Then combining the probability with the potential decrement mentioned above gives a drift in the potential of $\Omega(\frac{P(x)}{e \cdot m})$ as well.

Summarizing the above analysis, the drift in the potential is always $\Omega(\frac{P(x)}{e \cdot m})$.
Since the potential value cannot increase during the optimization process, and it can be upper bounded by $2m$ where $m = \theta(n^2)$, the Multiplicative Drift Theorem~\cite{doerr2012multiplicative} gives that the algorithm takes expected time $O(m \log n)$ to find the solution $x^{\dag}$ with $P(x^{\dag}) = 0$. 
\end{proof}
\end{theorem}

\subsection{GSEMO and Its Variants}

In the subsection, we separately study the performance of the GSEMO and its two variants with search points in edge-based representation to get a feasible solution to the 2H-(1,2)-MSTP on $G = (\{r\} \cup V,E,W)$.

\begin{theorem}
\label{thm:performance f_M}
The GSEMO with search points in edge-based representation takes expected time $O(mn \log n)$ to obtain a feasible solution to the 2H-(1,2)-MSTP on $G = (\{r\} \cup V,E,W)$.

\begin{proof}
Since the algorithm starts with a population that contains an arbitrary solution, we first consider the expected time of the algorithm to obtain a solution $x_0$ with $|x_0|_1 = 0$.
Let $S$ be the population maintained by the GSEMO, and $P_1 = \min_{x \in S} |x|_1$ be the first potential of the GSEMO. 
Observe that $P_1$ cannot increase during the optimization process by the dominance with respect to $\dominates_{\mathrm{GSEMO}}$.  
Let $x' = arg \min_{x \in S} |x|_1$. 
For the $|x'|_1$ 1-bits in $x'$, standard bit mutation selects exactly one of them and nothing else with probability $\Omega(\frac{|x'|_1}{e \cdot mn})$ (recall that the population size of the GSEMO is not greater than $n+1$), and its execution generates a solution $x''$ with Hamming weight $|x'|_1 -1$ that can be accepted by the algorithm (according to the dominance with respect to $\dominates_{\mathrm{GSEMO}}$). 
Thus the drift in the potential $P_1$ is $\Omega(\frac{|x'|_1}{e \cdot mn})$. 
Combining the drift with the fact that $P_1 \le m$, the Multiplicative Drift Theorem~\cite{doerr2012multiplicative} gives that the algorithm takes expected time $O(mn \log m) = O(mn \log n)$ to find a population with $P_1 = 0$.

Now we consider the process of the algorithm to obtain a feasible solution based on the solution $x_0$ (note that $x_0$ cannot be replaced by any solution according to the dominance with respect to $\dominates_{\mathrm{GSEMO}}$).
Let $S'$ be the subset of $S$ in which each solution $x$ satisfies $|x|_1 + N_{d>2}(x) = n$.
As $x_0$ satisfies $|x_0|_1 + N_{d>2}(x_0) = n$, $S'$ cannot be empty. 
Let $P_2 = n - \max_{x \in S'} |x|_1$ be the second potential of the GSEMO, and $x' = arg \max_{x \in S'} |x|_1$.
Observe that $P_2$ cannot increase during the optimization process by the dominance with respect to $\dominates_{\mathrm{GSEMO}}$, and $x'$ is a feasible solution if and only if $P_2 = 0$.
Thus in the following discussion, we assume that $|x'|_1 < n$, i.e., $N_{d>2}(x') = n - |x'|_1 > 0$.

For any vertex $v$ with $d_{G(x')}(v,r) >2$, flipping the 0-bit corresponding to the edge between $v$ and $r$ in $x'$ generates a new solution $x''$ with $N_{d>2}(x'') \le N_{d>2}(x')-1$ and $f(x'') \le f(x') + 2 -m^2$.
If $S$ has a solution $x_1$ with $|x_1|_1 = |x'|_1 + 1$, then $N_{d>2}(x_1) > N_{d>2}(x') -1$; otherwise, a contradiction to the definition of $x'$.
Thus $f'(x_1) > f(x'')$ (if $x_1$ exists), and the solution $x''$ can be accepted by the algorithm. 
That is, the potential is decreased by 1.
Considering the $N_{d>2}(x') = P_2$ many 0-bits in $x$ that correspond to the edges between $r$ and the vertices $v$ with $d_{G(x')}(v,r) >2$, standard bit mutation selects exactly one of them and nothing else with probability $\Omega(\frac{P_2}{e \cdot mn})$.
Thus the drift in the potential $P_2$ is $\Omega(\frac{P_2}{e \cdot mn})$.
Combining the drift with the fact that $P_2 \le n$, the Multiplicative Drift Theorem~\cite{doerr2012multiplicative} gives that the algorithm takes expected time $O(mn \log n)$ to find a population with $P_2 = 0$, after the acceptance of $x_0$.

Summarizing the above analysis gives that the algorithm takes expected time $O(mn \log n)$ to find a feasible solution. 
\end{proof}
\end{theorem}

\begin{figure*}[t]
  \centering     \footnotesize 
    \includegraphics[scale=0.3]{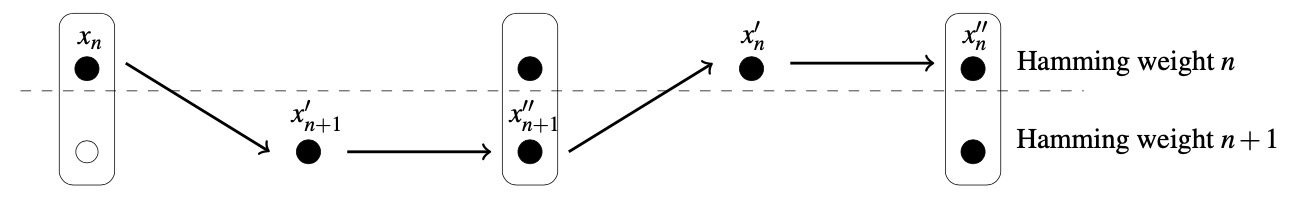}
    \vspace*{-1mm}
\caption{An illustration for the proof of Theorem~\ref{thm:performance f_M1}.
Each rectangle represents a population, and each solid circle represents a solution in the population (note that the hollow circle indicates that the population may or may not have such a solution). The solutions above and below the dashed line have Hamming weight $n$ and $n+1$, respectively.
}
\label{fig:illustration of GSEMO-1}
\end{figure*}

\begin{theorem}
\label{thm:performance f_M1}
The GSEMO-1 with search points in edge-based representation takes expected time $O(mn)$ to obtain a feasible solution to the 2H-(1,2)-MSTP on $G = (\{r\} \cup V,E,W)$.

\begin{proof}
We first consider the expected time of the algorithm to find a solution with Hamming weight $n$, using the potential function $P = \min_{x \in S} \left||x|_1 - n \right|$.
W.l.o.g., assume that the population $S$ maintained by the GSEMO-1 has no solution $x$ with $|x|_1 = n$.
Then the dominance with respect to $\dominates_{\textup{GSEMO-1}}$ indicates that $S$ has a unique solution $x_1$.

Case (1). $|x_1|_1 < n$. Observe that $G(x_1)$ has at least $n - |x_1|_1$ many connected components except $C_r(x_1)$, and $G$ has at least $n - |x_1|_1$ many edges between $r$ and them. 
Standard bit mutation selects exactly one of the 0-bits corresponding to the $n - |x_1|_1$ many edges in $x_1$ and nothing else with probability $\Omega(\frac{n - |x_1|_1}{e \cdot m})$ (recall that the population size of the algorithm can be upper bounded by 2), and its execution generates a new solution $x'_1$ with $|x'_1| = |x_1|_1 +1 \le n$ and $f(x'_1) < f(x_1)$ (as $N_{d > 2}(x'_1) < N_{d > 2}(x_1)$). 
Thus $x'_1 \dominates_{\textup{GSEMO-1}} x_1$, implying that $x'_1$ can be accepted by the algorithm, and the potential decreases by 1. 

Case (2). $|x_1|_1 > n$. Lemma~\ref{lem:update solution with weight greater than n} shows that there are $N_{\textup{cc}}(x_1) + |x_1|_1 - n -1$ many 1-bits in $x_1$, each of whose flip results in a new solution $x'_1$ with $c(x'_1) < c(x_1)$ and $N_{d>2}(x'_1) = N_{d>2}(x_1)$. 
If $x'_1$ has Hamming weight $n$, then no solution in $S$ is comparable to it with respect to $\dominates_{\textup{GSEMO-1}}$; otherwise, $x'_1$ dominates $x_1$ with respect to $\dominates_{\textup{GSEMO-1}}$ as $|x'_1| = |x_1| -1 \ge n$ and $f(x'_1) < f(x_1)$. 
Thus we have that $x'_1$ can be accepted by the algorithm, and the potential decreases by 1. 
Since $N_{\textup{cc}}(x_1) \ge 1$, $N_{\textup{cc}}(x_1) + |x_1|_1 - n -1 \ge |x_1|_1 - n$, and standard bit mutation selects exactly one of the 0-bits corresponding to the $N_{\textup{cc}}(x_1) + |x_1|_1 - n -1$ many edges in $x_1$ and nothing else with probability $\Omega(\frac{|x_1|_1 - n}{e \cdot m})$. 

The above discussion shows that the drift in the potential is always $\Omega (\frac{P}{e \cdot m})$. 
As $P \le m-n$, and its value cannot increase during the optimization process by the dominance with respect to $\dominates_{\textup{GSEMO-1}}$, the Multiplicative Drift Theorem~\cite{doerr2012multiplicative} gives that the algorithm takes expected time $O(m \log (m-n)) = O(m \log n)$ to obtain a population with potential $0$, which contains a solution with Hamming weight $n$. 
Note that before the acceptance of the solution with Hamming weight $n$, the population size remains 1.

Now we assume that the population $S$ contains a solution $x_n$ with $|x_n|_1 = n$ and consider the expected time of the algorithm to find a feasible solution based on $x_n$.
If $N_{d>2}(x_n) = 0$, then $x_n$ is feasible, and the proof is done.

Thus in the following discussion, $x_n$ is assumed to be infeasible, i.e., $N_{d >2}(x_n) > 0$.  
Let $v \in V$ be a vertex with $d_{G(x_n)}(v,r) > 2$, and $x'_{n+1}$ be the solution obtained by flipping the 0-bit corresponding to the edge $[v,r]$ in $x_n$ and nothing else. 
Observe that $N_{d>2}(x'_{n+1}) < N_{d>2}(x_{n})$ and $W([v,r]) \le 2$, thus 
$$f(x'_{n+1}) \le f(x_{n})+2 - m^2.$$ 
Standard bit mutation selects the 0-bit corresponding to the edge $[v,r]$ in $x_n$ and nothing else with probability $\Omega(1/m)$, thus the algorithm takes expected time $O(m)$ to get $x'_{n+1}$.

If the population $S$ has no solution with Hamming weight~$n+1$, or $f(x'_{n+1}) \leq f(x_{n+1})$ for the solution $x_{n+1}$ maintained in the population with Hamming weight $n+1$, then the solution $x'_{n+1}$ is accepted; otherwise, rejected. 
In the following discussion, we assume that the population contains a solution $x''_{n+1}$ with Hamming weight $n+1$ such that $f(x''_{n+1}) \leq f(x'_{n+1})$. 

By Lemma~\ref{lem:update solution with weight greater than n}, there exists a 1-bit in $x''_{n+1}$ whose flip results in a solution $x'_n$ with Hamming weight $n$ such that $N_{d>2}(x'_{n}) = N_{d>2}(x''_{n+1})$ and $f(x'_{n}) < f(x''_{n+1})$, where the execution of the corresponding mutation takes expected time $O(m)$. Moreover, for the solution $x'_n$, we have 
$$f(x'_{n}) < f(x''_{n+1}) \leq f(x'_{n+1}) \le f(x_{n})+2 - m^2.$$
Let $x''_n$ be the solution with Hamming weight $n$ in the population, after the construction of $x'_n$. For $x''_n$, it has $f(x''_{n}) \le f(x'_{n})$.

The update process of the solution with Hamming weight $n$ in the population is illustrated in Figure~\ref{fig:illustration of GSEMO-1}. 
Considering the mutation generating $x'_{n+1}$ and the one generating $x'_n$, the algorithm totally takes expected time $O(m)$ to get the improved solution $x''_n$ with $|x''_n|_1 = n$ and
$$f(x''_{n}) \le f(x_{n}) + 1 - m^2.$$

Combining the above conclusion with the fact that the value of $f(x_{n})$ can be upper bounded by $2n + m^2 n$, and that the size of the population can be upper bounded by 2, the Additive Drift Theorem~\cite{he2004study} gives that the algorithm takes expected time $O(mn)$ to get a solution $x^*$ with $|x^*|_1 = n$ and $f(x^*) < m^2$ (i.e., $x^*$ is a feasible solution) starting with the solution $x_n$. 
Apparently, $O(mn)$ is more expensive than the expected time $O(m \log n)$ to get $x_n$, thus we have the claimed expected time $O(mn)$ for the GSEMO-1.
\end{proof}
\end{theorem}

\begin{theorem}
\label{thm:performance f_M2}
The GSEMO-2 with search points in edge-based representation takes expected time $O(m \log n)$ to obtain a feasible solution to the 2H-(1,2)-MSTP on $G = (\{r\} \cup V,E,W)$.

\begin{proof}
We first analyze the expected time of the algorithm to find a solution $x^{\dag}$ with $N_{\textup{cc}}(x^{\dag}) =1$, using the potential function $P_1 = \min_{x \in S} (N_{\textup{cc}}(x) - 1)$. 
W.l.o.g., assume that the population $S$ maintained by the algorithm has no solution $x$ with $N_{\textup{cc}}(x) =1$. 
Then the dominance with respect to $\dominates_{\textup{GSEMO-2}}$ indicates that the population $S$ has a unique solution $x_1$.  

For a connected component $C$ in $G(x_1)$ except $C_r(x_1)$, as it has at least one vertex, there is at least one edge between $r$ and $C$, whose inclusion into $G(x_1)$ generates a new solution $x'_1$ with $N_{\textup{cc}}(x'_1) = N_{\textup{cc}}(x_1)-1$.
Although $|V_d(x'_1)|$ may be greater than $|V_d(x_1)|$, $|V_d(x'_1)| - |V_d(x_1)| \le n$, implying that $f_{\textup{M2}}^1(x'_1) \le f_{\textup{M2}}^1(x_1) - m^2 + n$, and $x'_1$ dominates $x_1$ with respect to $\dominates_{\textup{GSEMO-2}}$.  
Thus $x'_1$ can be accepted by the algorithm, and the potential $P_1$ is decreased by 1.
Considering the $N_{\textup{cc}}(x_1) -1$ many such edges between $r$ and the $N_{\textup{cc}}(x_1) -1$ many connected components in $G(x_1)$ except $C_r(x_1)$, standard bit mutation selects exactly one of them and nothing else with probability $\Omega(\frac{P_1}{e \cdot m})$, and the drift in the potential is $\Omega(\frac{P_1}{e \cdot m})$.

Combining the drift with the fact $P_1 \le n$ and that $P_1$ cannot increase during the optimization process by the dominance with respect to $\dominates_{\textup{GSEMO-2}}$, the Multiplicative Drift Theorem~\cite{doerr2012multiplicative} gives that the algorithm takes expected time $O(m \log n)$ to find a population with $P_1 = 0$, which contains a solution $x^{\dag}$ with $N_{\textup{cc}}(x^{\dag}) =1$. 

Now we analyze the expected time of the algorithm to find a solution $x^{\ddag}$ with $N_{\textup{cc}}(x^{\ddag}) =1$ and $|V_d(x^{\ddag})| = 0$ based on the solution $x^{\dag}$, using the potential function $P_2 = \min_{x \in S} |V_d(x)|$.
Note that after the acceptance of the solution $x^{\dag}$, no solution $x$ with $N_{\textup{cc}}(x) > 1$ can be accepted by the dominance with respect to $\dominates_{\textup{GSEMO-2}}$, thus all solutions $x$ considered in the following discussion are assumed to have $N_{\textup{cc}}(x) = 1$.
W.l.o.g., assume that the population $S$ maintained by the algorithm has no solution $x$ with $|V_d(x)| = 0$. 
Then the dominance with respect to $\dominates_{\textup{GSEMO-2}}$ indicates that the population $S$ has a unique solution $x_1$.  

For a vertex $v \in V_d(x_1)$, including the edge between $v$ and $r$ into $G(x_1)$ generates a new solution $x'_1$ with $N_{\textup{cc}}(x'_1) = N_{\textup{cc}}(x_1) = 1$ and $|V_d(x'_1)| \le |V_d(x_1)|-1$. 
Thus $x'_1$ also can be accepted by the algorithm, and the potential $P_2$ is decreased by at least one.
Considering all 0-bits in $x_1$ corresponding to the edges between $r$ and the vertices in $V_d(x_1)$, standard bit mutation selects exactly one of them and nothing else with probability $\Omega(\frac{P_2}{e \cdot m})$, and the drift in the potential $P_2$ is $\Omega(\frac{P_2}{e \cdot m})$. 

Combining the drift with the fact $P_2 \le n$ and that $P_2$ cannot increase during the optimization process by the dominance with respect to $\dominates_{\textup{GSEMO-2}}$, the Multiplicative Drift Theorem~\cite{doerr2012multiplicative} gives that after the acceptance of $x^{\dag}$, the algorithm takes expected time $O(m \log n)$ to find a population with $P_2 = 0$, which contains a solution $x^{\ddag}$ with $N_{\textup{cc}}(x^{\ddag}) =1$ and $|V_d(x^{\ddag})| = 0$. 

Observe that the solution $x^{\ddag}$ may contain cycles, i.e., $|x^{\ddag}|_1 > n$. Lemma~\ref{lem:update solution with weight greater than n} shows that there are $N_{\textup{cc}}(x^{\ddag}) + |x^{\ddag}|_1 - n-1$ 1-bits in $x^{\ddag}$, each of whose flip results in a solution $x'$ with $|x'|_1 < |x^{\ddag}|_1$, $c(x') < c(x^{\ddag})$, $N_{\textup{cc}}(x') = N_{\textup{cc}}(x^{\ddag}) = 1$, and $N_{d >2}(x') = N_{d >2}(x^{\ddag})  = 0$, implying that $f_{\textup{M2}}^1(x') = f_{\textup{M2}}^1(x^{\ddag}) = 1$,  $f_{\textup{M2}}^2(x') < f_{\textup{M2}}^2(x^{\ddag})$, and $x'$ can be accepted by the algorithm. 
Standard bit mutation selects exactly one of the 1-bits in $x$ corresponding to the $N_{\textup{cc}}(x^{\ddag}) + |x^{\ddag}|_1 - n-1 = |x^{\ddag}|_1 - n$ (as $N_{\textup{cc}}(x^{\ddag}) = 1$) edges and nothing else with probability $\Omega(\frac{|x^{\ddag}|_1 - n}{e \cdot m})$. 
Combining the probability with the fact $|x^{\ddag}|_1 - n \le m-n$, we can derive that the algorithm totally takes expected time 
$$O(\sum_{i=1}^{m-n} \frac{m}{i}) = O(m \log (m-n)) = O(m \log n),$$
to find a solution $x^*$ with $|x^*|_1 = n$ and $f_{\textup{M2}}^1(x^*) = 0$ starting with the solution $x^{\ddag}$.

Summarizing the above analysis, the algorithm takes expected time 
$O(m \log n)$ to obtain a feasible solution of $G$. 
\end{proof}
\end{theorem}

The above proof of Theorem~\ref{thm:performance f_M2} only considers the solution $x$ with $f_{\textup{M2}}^1(x) = 0$ in the population maintained by the GSEMO-2. For the other possible solution $x'$ with $f_{\textup{M2}}^1(x') = 1$ in the population, we will show its power to help the algorithm to find a solution with an improved ratio in the next section.

\section{Performance of the Four Algorithms with Search Points in Edge-based Representation for 3/2-Approximation}

The section studies the performance of the four algorithms to get an approximate solution with ratio 3/2 to the 2H-(1,2)-MSTP on $G = (\{r\} \cup V,E,W)$, based on their abilities to emulate local search operations.

\begin{figure*}[t]
  \centering     \footnotesize 
\includegraphics[scale=0.25]{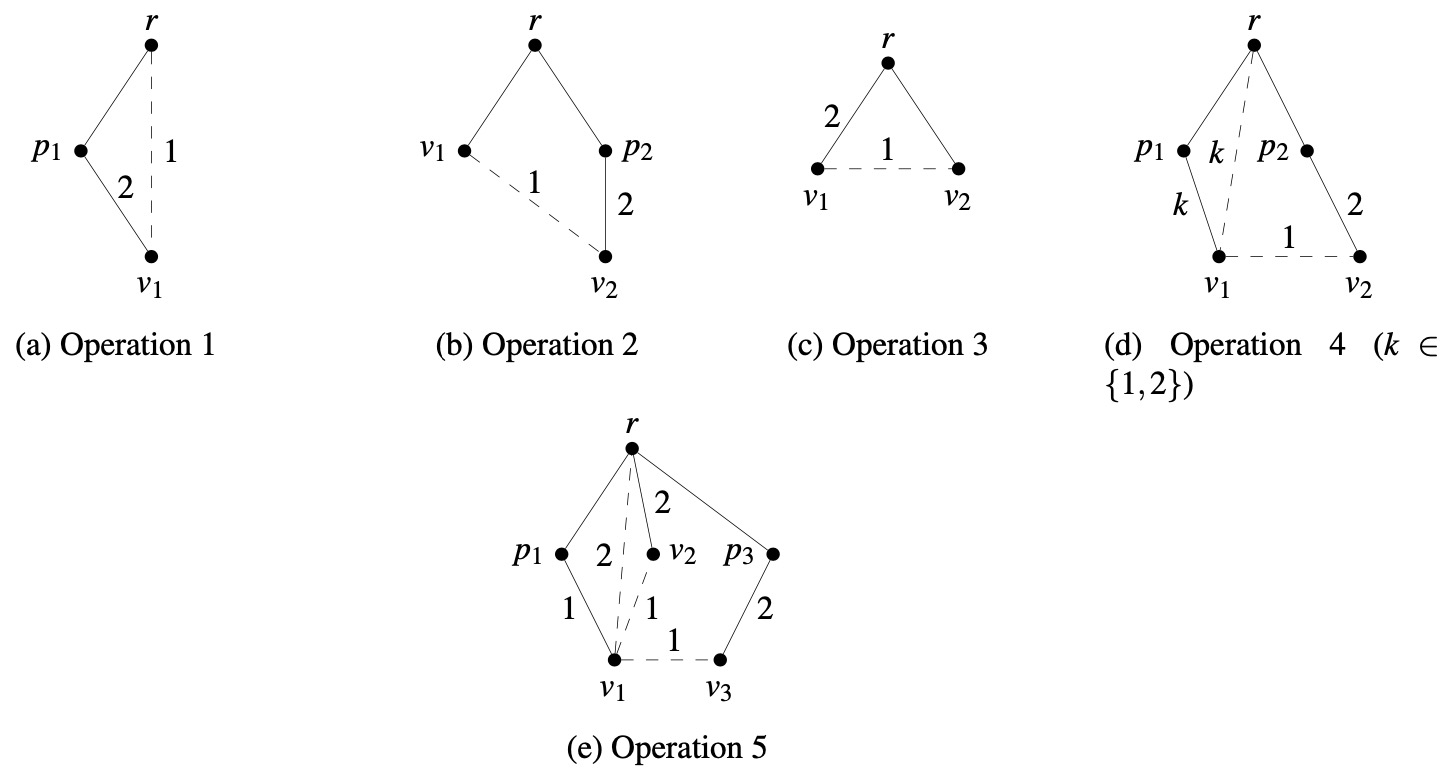}
\caption{Illustrations of Operation 1-5 considered in Theorem~\ref{thm:(1+1) EA 3/2}. In particular, the illustration of Operation 5 given above only considers the case that $v_2$ is a child of $r$ with no children, and $v_3$ is a grandchild of $r$ in $G(x_1)$.}
\label{fig:illustration of operations}
\end{figure*}

\begin{figure*}[t]
  \centering     \footnotesize 
\includegraphics[scale=0.25]{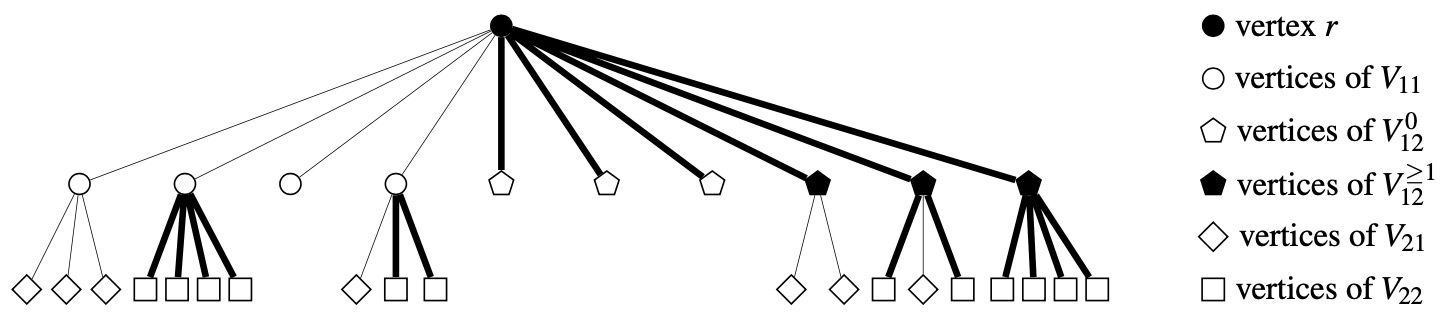}
\caption{An illustration for the partition of the vertices in $G(x_1)$. The thickness of edges is used to distinguish the weights on the edges: Thin ones have weight 1, and thick ones have weight 2. 
}
\label{fig:illustration of the partition}
\end{figure*}

\begin{theorem}
\label{thm:(1+1) EA 3/2}
The $(1{+}1)$ EA with search points in edge-based representation takes expected time $O(m^6 n)$ to obtain a 3/2-approximate solution to the 2H-(1,2)-MSTP on $G = (\{r\} \cup V,E,W)$.

\begin{proof}
Assume that the algorithm has obtained a feasible solution $x_1$. 
Theorem~\ref{thm:performance (1+1) EA 2} indicates that the process of the algorithm to get $x_1$ takes expected time $O(m \log n)$. 
In the following discussion, we introduce several operations to optimize $x_1$. 

We start with some related notions. 
As $G(x_1)$ is a spanning tree of $G$, if we treat the specific vertex $r$ as the root of the tree, then there is a well-defined ancestor-descendant relationship in $G(x_1)$. 
More specifically, for any edge $[v,v']$ with endpoints $v$ and $v'$ in $G(x_1)$, if $v$ is in the unique path connecting $v'$ and $r$ in $G(x_1)$, then $v$ is the {\it parent} of $v'$, and $v'$ is a {\it child} of $v$ in $G(x_1)$.
If $v'$ has a child $v''$ in $G(x_1)$, then $v''$ is a {\it grandchild} of $v$, and $v$ is the {\it grandparent} of $v''$ in $G(x_1)$.

{\bf Operation 1.} If there is a grandchild $v_1$ of $r$ in $G(x_1)$ such that the edge between $v_1$ and its parent $p_1$ in $G(x_1)$ has weight 2, but $W([v_1,r]) = 1$, then swap the edge $[v_1,p_1]$ with the edge $[v_1,r]$. The illustrations of Operation 1 and the following four operations are given in Figure~\ref{fig:illustration of operations}.

{\bf Operation 2.} If there is a child $v_1$ and a grandchild $v_2$ of $r$ in $G(x_1)$ such that the edge between $v_2$ and its parent $p_2$ (not $v_1$) has weight 2, but $W([v_1,v_2]) = 1$, then swap the edge $[v_2,p_2]$ with the edge $[v_1,v_2]$.

{\bf Operation 3.} If there are two children $v_1$ and $v_2$ of $r$ in $G(x_1)$ such that $v_1$ has no child, and $W([v_1,r])=2 > W([v_1,v_2])=1$, then swap the edge $[v_1,r]$ with the edge $[v_1,v_2]$. 

Obviously, each application of Operation 1, Operation 2, and Operation 3 on $x_1$ gets an improved solution $x'_1$ with $c(x'_1) = c(x_1) -1$ that can be accepted by the algorithm.
Since the mutation corresponding to an application of them can be generated with probability $\Omega(1/m^2)$, the algorithm takes expected time $O(m^2)$ to get the improved solution $x'_1$.
 
{\bf Operation 4.} If there is a grandchild $v_1$ of $r$ and a vertex $v_2$ that is either a grandchild or a child with no child of $r$ in $G(x_1)$, such that the edge between $v_1$ and its parent $p_1$ in $G(x_1)$ and the edge $[v_1,r]$ have the same weight, but the edge between $v_2$ and its parent $p_2$ (may be $r$) in $G(x_1)$ has a larger weight than the edge $[v_1,v_2]$, then swap the edges $[v_1,p_1]$ and $[v_2,p_2]$ with the edges $[v_1,r]$ and $[v_1,v_2]$.

Each application of Operation 4 on $x_1$ gets an improved solution $x'_1$ with $c(x'_1) = c(x_1) -1$, which can be accepted by the algorithm.
The mutation corresponding to the application can be generated with probability $\Omega(1/m^4)$, thus the algorithm takes expected time $O(m^4)$ to get $x'_1$.
Now we consider a grandchild $v_1$ of $r$ in $G(x_1)$ and two vertices $v_2$ and $v_3$, each of which is either a grandchild  of $r$ or a child of $r$ with no child in $G(x_1)$. 
Let $p_1$, $p_2$, and $p_3$ be the parents of $v_1$, $v_2$, and $v_3$, respectively ($p_2$ and $p_3$ may be $r$ if $v_2$ and $v_3$ are the children of $r$ in $G(x_1)$).

{\bf Operation 5.} If $W([v_1,p_1]) = 1 = W([v_1,v_2]) = W([v_1,v_3])$, but $W([v_1,r]) = 2 = W([v_2,p_2]) = W([v_3,p_3])$, then swap $[v_1,p_1]$, $[v_2,p_2]$, $[v_3,p_3]$ with $[v_1,r]$, $[v_1,v_2]$, and $[v_1,v_3]$.

Each application of Operation 5 gets an improved solution $x'_1$ with $c(x'_1) = c(x_1) -1$ that can be accepted by the algorithm. 
The mutation corresponding to the application can be generated with probability $\Omega(1/m^6)$, thus the algorithm takes expected time $O(m^6)$ to get $x'_1$.

Summarizing the analysis for the above operations, each application improve the fitness of the maintained solution by 1. Combining the conclusion with the fact that $c(x_1) \leq 2n$ and $c(x^*) \geq n$, where $x^*$ is an optimal solution to the 2H-(1,2)-MSTP on $G$, we can derive that Operation 1-5 can be applied at most $n$ times.
Thus, starting with $x_1$, the algorithm takes expected time $O(m^6 n)$ to get a feasible solution $x_2$ on which Operation 1-5 are not applicable.

Now we analyze the cost of the solution $x_2$.
We start with some related notations. Given a feasible solution $x$, the vertices of $V$ can be partitioned into the following subsets according to the structure of $G(x)$ (an illustration can be found at Figure~\ref{fig:illustration of the partition}).
\begin{enumerate}[1)] 
\item $V_{11}(x)$, contains all the vertices $v \in V$, where $v$ is a child of $r$ in $G(x)$, and $W([v,r]) = 1$; 
\item $V_{12}(x)$, contains all the vertices $v \in V$, where $v$ is a child of $r$ in $G(x)$, and $W([v,r]) = 2$; 
\item $V_{21}(x)$, contains all the vertices $v \in V$, where $v$ is a grandchild of $r$ in $G(x)$, and $W([v,p]) = 1$ ($p$ is the parent of $v$ in $G(x)$); 
\item $V_{22}(x)$, contains all the vertices $v \in V$, where $v$ is a grandchild of $r$ in $G(x)$, and $W([v,p]) = 2$ ($p$ is the parent of $v$ in $G(x)$).
\end{enumerate}
Moreover, the vertices in $V_{12}(x)$ are partitioned into the following two subsets.
\begin{enumerate}[1)] 
\item $V_{12}^0(x)$, contains all the vertices $v \in V_{12}(x)$, where $v$ has no child in $G(x)$; 
\item $V_{12}^{\ge 1}(x)$, contains all the vertices $v \in V_{12}(x)$, where $v$ has at least one child in $G(x)$. 
\end{enumerate}

Consider a vertex $v \in V_{22}(x_2) \cup V_{12}^0(x_2)$.
Firstly, $W([v,v']) = 2$ for any vertex $v' \in \{r\} \cup V_{11}(x_2) \cup V_{12}(x_2) \setminus \{v\}$; otherwise, Operation 1 or 2 or 3 is applicable on $x_2$.
Then $W([v,v']) = 2$ for any vertex $v' \in V_{22}(x_2) \setminus \{v\}$; otherwise, Operation 4 is applicable on $x_2$.
Thus, $W([v,v']) = 2$ for any vertex $v' \in \{r\} \cup V_{11}(x_2) \cup V_{12}(x_2) \cup V_{22}(x_2) \setminus \{v\}$.
If there exists a vertex $v' \in V_{21}(x_2)$ with $W([v,v']) = 1$, then $W([v',r]) = 2$; otherwise, Operation 4 is applicable on $x_2$.

Let $v_1$ be a vertex of $V_{22}(x_2) \cup V_{12}^0(x_2)$ with $N_1(v_1) \neq \O$ (recall that $N_1(v_1)$ is the set containing all the vertices $v'$ in $G$ with $W([v_1,v']) = 1$). 
The above analysis shows that $N_1(v_1) \subset V_{21}(x_2)$). 
Now we consider the possible cases for the vertex $v_1$ in $G(x^*)$, where $x^*$ is an optimal solution to the 2H-(1,2)-MSTP on $G$.
If the parent $p_1$ of $v_1$ in $G(x^*)$ is a vertex of $\{r\} \cup V \setminus N_1(v_1)$, then the edge between $v_1$ and $p_1$ in $G(x^*)$ has weight 2; otherwise (i.e., $p_1 \neq r$), the above analysis shows that the edge between $p_1$ and its parent $r$ in $G(x^*)$ has weight~2.

Assume that the parent $p_1$ of $v_1$ in $G(x^*)$ is a vertex of $N_1(v_1)$, where $N_1(v_1) \subset V_{21}(x_2)$. 
We have that if there is a vertex $v_2$ in $V_{22}(x_2) \cup V_{12}^0(x_2) \setminus \{v_1\}$ that is also the child of $p_1$ in $G(x^*)$, then $W([p_1,v_2]) = 2$; otherwise, Operation 5 is applicable on $G(x_2)$ with respect to $v_1$, $v_2$, and $p_1$.   
That is, there is no vertex $v_2$ in $V_{22}(x_2) \cup V_{12}^0(x_2) \setminus \{v_1\}$ with $N_1(v_1) \cap N_1(v_2) \neq \O$.
In other words, if there is a subset $V' \subset V_{22}(x_2) \cup V_{12}^0(x_2)$ in which the vertices have the same common parent $p$ in $G(x^*)$, then one of the following three cases holds: 

\smallskip
\noindent {\bf Case (I).} $p$ is the vertex $r$, thus all the edges between $p$ and the vertices in $V'$ have weight 2;\\ 
\noindent {\bf Case (II).} $p$ is a child of $r$, and there is exactly one vertex $v \in V'$ with $v \in N_1(p)$. 
That is, all edges between $p$ and the vertices in $V' \setminus \{v\}$ have weight 2, the edge $[v,p]$ has weight 1, and the edge $[p,r]$ has weight 2;\\
\noindent {\bf Case (III).} $p$ is a child of $r$, and there is no vertex $v \in V'$ with $p \in N_1(v)$. That is, all edges between $p$ and the vertices in $V'$ have weight 2.
\smallskip

Summarizing the above analysis, we have 
\begin{eqnarray*}
c(x^*) &\geq& n+ |V_{12}^0(x_2)| + |V_{22}(x_2)|.
\end{eqnarray*}
The following inequality can be easily derived, where the last inequality relation holds because each vertex in $V^{\ge 1}_{12}(x_2)$ has at least one child in $G(x_2)$, i.e., $|V^{\ge 1}_{12}(x_2)| \leq n/2$. 
\begin{eqnarray*}
c(x_2) &=& |V_{11}(x_2)| + 2|V_{12}(x_2)|+|V_{21}(x_2)| +2|V_{22}(x_2)| \\
&=& n + |V_{12}(x_2)| + |V_{22}(x_2)|  \\
&=& n + |V^0_{12}(x_2)|+ |V^{\ge 1}_{12}(x_2)| + |V_{22}(x_2)| \\
&\leq& 3n/2 + |V^0_{12}(x_2)| + |V_{22}(x_2)|.
\end{eqnarray*}
 
Therefore, the approximation ratio of $x_2$ is
\begin{eqnarray*}
\frac{c(x_2)}{c(x^*)} \leq \frac{3n/2 + |V^0_{12}(x_2)| + |V_{22}(x_2)|}{n+ |V^0_{12}(x_2)| + |V_{22}(x_2)|} \leq \frac{3}{2} . 
\end{eqnarray*}

The above conclusion gives that the algorithm takes expected time $O(m^6 n)$ to obtain an approximate solution with ratio 3/2.   
\end{proof}
\end{theorem}

Using almost the same reasoning given in the proof for Theorem~\ref{thm:(1+1) EA 3/2} and the population sizes of the GSEMO and GSEMO-1, we can get the following theorem.

\begin{theorem}
\label{thm:improved performance f_M}
The GSEMO and GSEMO-1 with search points in edge-based representation take expected time $O(m^6 n^2)$ and $O(m^6 n)$, respectively, to obtain a 3/2-approximate solution to the 2H-(1,2)-MSTP on $G = (\{r\} \cup V,E,W)$.
\end{theorem}
 
Now we consider the performance of the GSEMO-2 to obtain a 3/2-approximate solution, based on the first four local search operations given in the proof for Theorem~\ref{thm:(1+1) EA 3/2} and two new local search operations.

\begin{figure*}[t]
  \centering     \footnotesize 
    \includegraphics[scale=0.3]{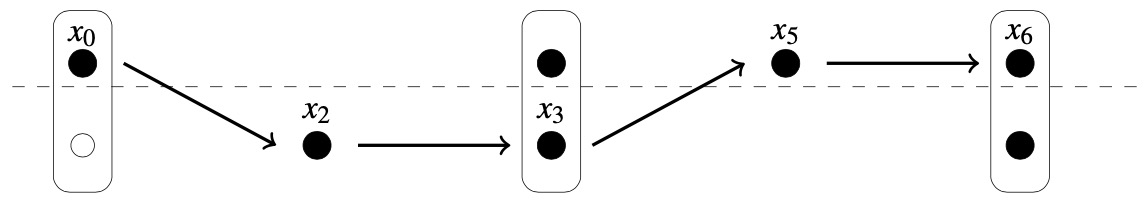}
\caption{An illustration for the proof of Theorem~\ref{thm:improved performance f_M2}.
Each rectangle represents a population, and each solid circle represents a solution in the population (note that the hollow circle indicates that the population may or may not have such a solution). For each solution $x$, if it locates above the dashed line, then $|V_d(x)| = 0$; otherwise, $|V_d(x)| = 1$.
}
\label{fig:illustration of GSEMO-2}
\end{figure*}

\begin{theorem}
\label{thm:improved performance f_M2}
The GSEMO-2 with edge-based representation takes expected time $O(m^4 n)$ to obtain a 3/2-approximate solution to the 2H-(1,2)-MSTP on $G = (\{r\} \cup V,E,W)$.

\begin{proof}
Assume that the algorithm has obtained a population that contains a feasible solution $x_0$, i.e., $N_{d > 2}(x_0) = 0$ and $|x_0|_1 = n$. 
Theorem~\ref{thm:performance f_M2} shows that the algorithm takes expected time $O(m \log n)$ to get such a population. 
In the following discussion, several local search operations are introduced to optimize $x_0$, including Operation 1-4 given in the proof for Theorem~\ref{thm:(1+1) EA 3/2} and two new operations given below.
Consider a grandchild $v_1$ of $r$ and two vertices $v_2$ and $v_3$, each of which is either a grandchild of $r$ or a child of $r$ with no child in $G(x_0)$. 
Let $p_2$ and $p_3$ be the parents of $v_2$ and $v_3$, respectively ($p_2$ and $p_3$ may be $r$ if $v_2$ and $v_3$ are the children of $r$ in $G(x_0)$).

{\bf Operation 6.} If $W([v_1,v_2]) = W([v_1,v_3]) = 1$, $W([v_2,p_2]) = W([v_3,p_3]) = 2$, and the population has no solution except $x_0$ or the other solution $x_1$ with $f_{\textup{M2}}^1(x_1) = 1$ in the population has $f_{\textup{M2}}^2(x_1) \ge f_{\textup{M2}}^2(x_0) -1$, then swap the edges $[v_2,p_2]$ and $[v_3,p_3]$ in $G(x_0)$ with $[v_1,v_2]$ and $[v_1,v_3]$. 

If Operation 6 is applicable, then the obtained solution $x_2$ has $V_d(x_2) = \{v_1\}$, $|x_2|_1 = |x_0|_1 = n$, $N_{\textup{cc}}(x_2) = 1$, $c(x_2) = c(x_0) - 2$ and $f_{\textup{M2}}^2(x_2) = f_{\textup{M2}}^2(x_0)-2$, implying that $x_2$ can be accepted by the algorithm.
Thus after the trial of Operation 6, the population contains a solution $x_3$ with $f_{\textup{M2}}^1(x_3) = 1$ and $f_{\textup{M2}}^2(x_3) \leq f_{\textup{M2}}^2(x_0) - 2$.
As $N_{\textup{cc}}(x_3) = 1$ and $f_{\textup{M2}}^2(x_3) \leq f_{\textup{M2}}^2(x_0) - 2$, $|x_3|_1 = n$ and $c(x_3) \le c(x_0) -2$.
Denote by $v$ the unique vertex in $V_d(x_3)$, and $p$ the parent of $v$ in $G(x_3)$. 

{\bf Operation 7.} If $f_{\textup{M2}}^2(x_3) \leq f_{\textup{M2}}^2(x_4) - 2$, where $x_4$ is the solution in the population with $f_{\textup{M2}}^1(x_4) = 0$, then swap the edge $[v,p]$ in $G(x_3)$ with the edge $[v,r]$.

Note that $x_4$ may be different from the above mentioned solution $x_0$, as during the phase that after the trial of Operation 6 but before the trial of Operation 7, $x_0$ may be replaced by other solutions. 
Thus $|x_4|_1 = n$ and $f_{\textup{M2}}^2(x_4) = c(x_4) \le c(x_0)$.
If Operation 7 is applicable, then the obtained solution $x_5$ has $|x_5|_1 = n$, $|V_d(x_5)| = 0$ and $N_{\textup{cc}}(x_5) = 1$, i.e., $x_5$ is a feasible solution.  
As the edges $[v,p]$ and $[v,r]$ may have weights 1 and 2, respectively, $c(x_5) \leq c(x_3) - 1 + 2 = c(x_3) + 1$. 
Therefore, 
$$c(x_5) \leq c(x_3) + 1 \le c(x_4) - 1 \le c(x_0) - 1.$$
That is, if Operation 7 is applicable, then the obtained solution $x_5$ can be accepted by the algorithm, replacing the maintained feasible solution $x_4$; otherwise, $f_{\textup{M2}}^2(x_3) > f_{\textup{M2}}^2(x_4) - 2$, implying $c(x_4) < c(x_0)$. 
Thus after the trial of Operation 7, the population contains a feasible solution $x_6$ with $c(x_6) < c(x_0)$.

Summarizing the above analysis, the trials of Operation 6 and 7 can emulate Operation 5 given in the proof for Theorem~\ref{thm:(1+1) EA 3/2} and improve the cost of the feasible solution in the population by at least 1. 
The update process of the solutions is given in Figure~\ref{fig:illustration of GSEMO-2}. 
The mutations corresponding to Operation 6 and 7 can be generated with probability $\Omega(1/m^4)$ and $\Omega(1/m^2)$, respectively, i.e., the algorithm totally takes expected time $O(m^4)$. 

Since $c(x_0) \leq 2 n$ and $c(x^*) \geq n$, where $x^*$ is an optimal solution to the 2H-(1,2)-MSTP on $G$, Operation 1-4 and 6-7 can be applied at most $O(n)$ times.
That is, starting with $x_0$, the algorithm takes expected time $O(m^4 n)$ to get a feasible solution $x_7$ on which Operation 1-4 and 6-7 are not applicable.
The reasoning given in the proof for Theorem~\ref{thm:(1+1) EA 3/2} implies that $x_7$ has approximation ratio 3/2.
\end{proof}
\end{theorem}

By the proof for Theorem~\ref{thm:(1+1) EA 3/2}, it is not hard to see that the traditional local search algorithm (more specifically, the algorithm has to consider all the edges involved in each local search operation simultaneously) takes time $O(m^6 n)$ to get a 3/2-approximate solution to the 2H-(1,2)-MSTP. 
Thus the GSEMO-2 has a better upper bound on the time complexity than the traditional local search algorithm (note that the time complexity of the GSEMO-2 is expected), and the mechanism to tolerate some infeasible solutions in the population as intermediates to accelerate the emulation of local search operations is efficient.

\section{Performance of the \oneone with Search Points in Vertex-based Representation for 3/2-Approximation}

The section considers the (1+1) EA (given in Algorithm~\ref{alg:1+1}) with search points in vertex-based representation and studies its performance with respect to a trivial fitness function $f_{\textup{vr}}(x) = c(x)$ (as all search points in vertex-based representation are feasible, it is unnecessary to penalize their infeasibility). 
Recall that a search point $x$ in vertex-based representation is represented as a bit string with length $n$, thus the probability $1/m$ given in Step 4 of the \oneone (see Algorithm~\ref{alg:1+1}) should be replaced with $1/n$.

\begin{theorem}
\label{thm:(1+1) EA 3/2 vertex-based representation}
The $(1{+}1)$ EA with search points in vertex-based representation takes expected time $O(n^4)$ to obtain a 3/2-approximate solution to the 2H-(1,2)-MSTP on $G = (\{r\} \cup V,E,W)$.

\begin{proof}
The proof for Theorem~\ref{thm:(1+1) EA 3/2} shows that if Operations 1-5 are not applicable on a 2-hop spanning tree, then it is a 3/2-approximate solution to the 2H-(1,2)-MSTP on $G = (\{r\} \cup V,E,W)$.
Thus in the following discussion, we analyze the expected time of the $(1{+}1)$ EA with search points in vertex-based representation to emulate the five local search operations. 

Let $x$ be the solution maintained by the algorithm. 
Recall that the vertices of $V \setminus V(x)$ are connected to the vertices of $V(x)$ with the minimum cost in $G(x)$, thus the algorithm emulates the five operations by adjusting the vertices chosen by $x$. 
The case indicated by Operation 1 (see the illustration given in Figure~\ref{fig:illustration of operations}(a)) shows that $v_1 \notin V(x)$ but $W([v_1,p_1]) = 2 > W([v_1,r]) = 1$. Thus Operation 1 can be emulated by flipping the bit corresponding to $v_1$ in $x$ from 0 to 1, and the algorithm takes expected time $O(n)$ to generate the mutation. 
Similar analysis can be applied to Operation 3, and the algorithm takes expected time $O(n)$ to emulate it. 

The case indicated by Operation 2 (see the illustration given in Figure~\ref{fig:illustration of operations}(b)) shows that $v_1 \in V(x)$ and $v_2 \notin V(x)$. Thus $v_2$ should be connected to a vertex of $V(x)$ with the minimum cost in $G(x)$.
However, the condition of Operation 2 that $W([v_2,v_1]) = 1 < W([v_2,p_2]) = 2$ indicates that $v_2$ is not connected in the optimal way, a contradiction.
Therefore, the case indicated by Operation 2 does not exist in $G(x)$.

For Operation 4 (see the illustration given in Figure~\ref{fig:illustration of operations}(d)), if $v_2$ is a child of $r$ in $G(x)$, i.e., $v_2 \in V(x)$, then it can be emulated by flipping the bit corresponding to $v_1$ from 0 to 1 and the bit corresponding to $v_2$ from 1 to 0; if $v_2$ is a grandchild of $r$ in $G(x)$, i.e., $v_2 \notin V(x)$, then it can be emulated by flipping the bit corresponding to $v_1$ from 0 to 1. 
Considering the above two cases for Operation 4, the algorithm takes expected time $O(n^2)$ to emulate it.

For Operation 5 (see the illustration given in Figure~\ref{fig:illustration of operations}(e)), if $v_2$ and $v_3$ are the children of $r$ in $G(x)$, then it can be emulated by flipping the bit corresponding to $v_1$ from 0 to 1 and the bits corresponding to $v_2$ and $v_3$ from 1 to 0; if exactly one of $v_2$ and $v_3$ (say $v_2$) is the child of $r$ in $G(x)$, then it can be emulated by flipping the bit corresponding to $v_1$ from 0 to 1 and the bit corresponding to $v_2$ from 1 to 0; if neither $v_2$ nor $v_3$ is the child of $r$ in $G(x)$, then it can be emulated by flipping the bit corresponding to $v_1$ from 0 to 1.
Considering the above three cases for Operation 5, the algorithm takes expected time $O(n^3)$ to emulate it.

Each application of Operations 1-5 improves the fitness of the solution by at least one. 
As $c(x) \le 2n$ and $c(x^*) \ge n$, where $x^*$ is an optimal solution to the 2H-(1,2)-MSTP on $G = (\{r\} \cup V,E,W)$, the process of the algorithm to obtain a solution on which Operations 1-5 are not applicable contains at most $n$ many applications of the five operations. 
Combining the above conclusion with the expected time of the algorithm to emulate them, we can derive that the process takes expected runtime $O(n^4)$, i.e., the algorithm obtains a 3/2-approximate solution to the problem within expected time $O(n^4)$. 
\end{proof}
\end{theorem}

\section{Conclusion}
\label{sec:conclusion}

In the paper we studied a constrained version of the Minimum Spanning Tree problem, named 2-Hop (1,2)-Minimum Spanning Tree problem (abbr. 2H-(1,2)-MSTP), on a complete edge-weighted graph in which each edge has weight 1 or 2. 
As the 2H-(1,2)-MSTP is NP-hard, we investigated the expected time of several evolutionary algorithms designed for it to obtain an approximate solution to the problem with a target ratio.
Two representations for the search points were considered: Edge-based representation and vertex-based representation. 

For the edge-based representation, a solution represented in this way may be infeasible, so firstly, we investigated the expected time of the (1+1) EA and GSEMO to obtain a feasible solution to the problem.
Observed that the large population of the GSEMO slows its optimization process, hence we presented its two variants (namely, GSEMO-1 and GSEMO-2), where each of them maintains a population with at most two solutions (a feasible one and an infeasible one).
The analysis for the performance of the two variants showed that the interplay between the two solutions can accelerate the optimization process.   
Secondly, by introducing several local search operations, we investigated the performance of the four algorithms mentioned above to get an approximate solution with ratio 3/2 to the 2H-(1,2)-MSTP. 
By comparing the expected time of the (1+1) EA and GSEMO-2 for 3/2 approximations, it is easy to see that the mechanism of the GSEMO-2 keeping a feasible solution and an infeasible one in the population has an advantage over the classic local search technique. That is because the local search operation with the worst expected time can be decomposed into two operations under the mechanism of the GSEMO-2, moreover, the two operations do not required to be accomplished at the same time.

For the vertex-based representation, the considered algorithm \oneone only needs to select the vertices that are neighbor to $r$, as all the other vertices are assumed to be connected to the neighbors of $r$ with the minimum cost. 
Thus the local search operations mentioned above that swap edges can be reformulated as the local search operations that swap vertices. 
Moreover, the number of vertices participating in a local search operation is less than that of the participating edges. 
Thus the expected time of the \oneone with search points in vertex-based representation to emulate these local search operations can be improved significantly.
Consequently, the \oneone with search points in vertex-based representation is shown to have better performance to obtain an approximate solution with ratio 3/2 to the 2H-(1,2)-MSTP, compared with the \oneone with search points in edge-based representation.

It is not hard to see that the reasoning given in the paper can be adapted to the 2H-(1,$\alpha$)-MSTP on a complete edge-weighted graph, in which each edge has weight 1 or $\alpha$ ($\alpha$ is an integer greater than 2).  
Moreover, the ideas (in particular, the mechanism of the GSEMO-1 and GSEMO-2 keeping a feasible solution and an infeasible solution) and reasoning introduced in the paper may be applied to design and analyze the evolutionary algorithms for the Bounded Diameter MSTP, the Uncapacitated Facility problem, the Cluster Median problem, and the {\it generalized} versions of the 2H-(1,$\alpha$)-MSTP, where the constraint on the number of hops is relaxed to an integer $\beta > 2$, or the weight on each edge in the input graph has more than two options.
Future work on these problems would be interesting and enrich the theoretical results for the behaviors of the evolutionary computing.

\bibliography{myreferences}

\end{document}